\documentclass[twoside,11pt]{article}

\usepackage[font=small]{caption}
\usepackage{tabulary}
\usepackage{amsmath}
\usepackage{calc}
\usepackage{bm}
\usepackage{lipsum}
\usepackage{booktabs}
\usepackage{url}
\usepackage{tablefootnote}
\usepackage[detect-all]{siunitx}
\usepackage[caption=false,font=footnotesize]{subfig}
\usepackage{jmlr2e}

\renewcommand{\ll}{{\lambda}}
\newcommand{\Lagr}{\mathcal{L}}
\newcommand{\xialpha}{{\xi\text{-}\alpha}}
\newcommand{\Lp}{{\Lambda_p}}
\newcommand{\LOOError}{{R_{\text{\tiny LOO}}}}
\newcommand{\T}{{T}}
\newcommand{\x}{{\mathbf{x}}}
\newcommand{\w}{{\mathbf{w}}}
\renewcommand{\a}{{\alpha}}
\newcommand{\ba}{{\bm{\alpha}}}
\newcommand{\sumL}{{\sum\limits_{i=1}^l}}
\newcommand{\sumLij}{{\sum\limits_{i,j=1}^l}}
\newcommand{\aV}{{\boldsymbol\alpha}}

\newcommand{\sumUBNoP}{{\sum\limits_{i=1,i \neq p }^{n^{*}}}}

\newcommand{\sumUBNoPpos}{{\sum \limits_{\substack{i=1,i \neq p, \\ y_i = y_p }}^{n^{*}}}}

\newcommand{\sumUBneg}{{\sum \limits_{\substack{i=1, \\ y_i \neq y_p}}^{n^{*}}}}
\newcommand{\sumUBpos}{{\sum \limits_{\substack{i=1, \\ y_i = y_p }}^{n^{*}}}}
\newcommand{\sumUB}{{\sum\limits_{i=1}^{n^{*}}}}

\newcommand{\sumB}{{\sum\limits_{i=n^{*}+1}^{n}}}
\newcommand{\sumBNoP}{{\sum\limits_{i=n^{*}+1, i \neq p}^{n}}}
\newcommand{\sumN}{{\sum\limits_{i=1}^n}}

\newcommand{\D}{{\Delta}}
\newcommand{\lV}{{\boldsymbol\lambda}}

\DeclareMathOperator*{\argmin}{argmin}

\jmlrheading{}{}{}{}{}{Ioannis Sarafis, Christos Diou and Anastasios
  Delopoulos}

\ShortHeadings{Span error bound for weighted SVM}{Sarafis, Diou and
  Delopoulos}

\firstpageno{1}

\begin{document}

\title{Span error bound for weighted SVM with applications in
  hyperparameter selection}

\author{\name Ioannis Sarafis \email sarafis@mug.ee.auth.gr 
  \AND Christos Diou \email diou@mug.ee.auth.gr
  \AND Anastasios Delopoulos \email adelo@eng.auth.gr \\
  \addr Multimedia Understanding Group \\ Electrical and Computer
  Engineering Department\\ Aristotle University of Thessaloniki,
  Greece 
}

\editor{}

\maketitle

\begin{abstract}
Weighted SVM (or fuzzy SVM) is the most widely used SVM variant owning
its effectiveness to the use of instance weights. Proper selection of
the instance weights can lead to increased generalization
performance. In this work, we extend the span error bound theory to
weighted SVM and we introduce effective hyperparameter selection
methods for the weighted SVM algorithm. The significance of the
presented work is that enables the application of span bound and
span-rule with weighted SVM. The span bound is an upper bound of the
leave-one-out error that can be calculated using a single trained SVM
model. This is important since leave-one-out error is an almost
unbiased estimator of the test error. Similarly, the span-rule gives
the actual value of the leave-one-out error. Thus, one can apply span
bound and span-rule as computationally lightweight alternatives of
leave-one-out procedure for hyperparameter selection. The main
theoretical contributions are: (a) we prove the necessary and
sufficient condition for the existence of the span of a support vector
in weighted SVM; and (b) we prove the extension of span bound and
span-rule to weighted SVM. We experimentally evaluate the span bound
and the span-rule for hyperparameter selection and we compare them
with other methods that are applicable to weighted SVM: the $K$-fold
cross-validation and the ${\xi}-{\alpha}$ bound. Experiments on 14
benchmark data sets and data sets with importance scores for the
training instances show that: (a) the condition for the existence of
span in weighted SVM is satisfied almost always; (b) the span-rule is
the most effective method for weighted SVM hyperparameter selection;
(c) the span-rule is the best predictor of the test error in the mean
square error sense; and (d) the span-rule is efficient and, for
certain problems, it can be calculated faster than $K$-fold
cross-validation.


\end{abstract}

\newpage

\section{Introduction}

Weighted SVM (or fuzzy SVM) uses importance weights, $C_i$, for each
training instance, $(\x_i, y_i)$, and is one of the most commonly used
variants of the soft-margin SVM \citep{lin2002fuzzy}. The algorithm
was initially proposed as a robust SVM-based alternative for problems
with outliers
\citep{lin2002fuzzy,lin2004training,wu2004incorporating}. The weights
regularize the misclassification penalty and therefore alter the
contribution of each instance in the final solution. Additionally,
\cite{lapin2014learning} showed that the instance weights of weighted
SVM can express the same type of prior knowledge that can be encoded
using privileged features, like the ones of the SVM+ algorithm
\citep{Vapnik2009SVM+}. The weighted SVM formulation is presented in
detail in Section \ref{subsec:weighted_svm_prelim}.

Weighted SVM can be applied on training instances with different
importance or certainty. The different importance may be due to
imbalanced or noisy classes, instances with feature noise (for
example, measurement noise, outliers), instances with label noise
(i.e. erroneous class label assignments), or instances with different
importance (for example, importance scores assigned by an expert).

Weighted SVM applications have used instance weights to reflect: the
importance of an instance using the class labels of its neighbors or
its closest class cluster \citep{Cheng2016,Wu2014}; the class
importance in problems with overlapping or imprecise label assignments
\citep{han2008fuzzy}; or the density of the class around the training
instance calculated with non-parametric methods
\citep{Bicego2009}. Also, the weights can be used to incorporate
problem-specific knowledge for the individual importance of the
training instances, for example: the score for bankruptcy in economics
applications \citep{Chaudhuri2011}; the importance of the instances
according to the user feedback in multimedia retrieval systems
\citep{Barrett2009}; or the instance importance in automatically
generated training sets from noisy sources, such as user clickthrough
data harvested from search engine logs \citep{sarafis2015building,
  sarafis2016online} or user-assigned tags
\citep{papapanagiotou2015improving}.

The selection of the SVM hyperparameters---including the
regularization parameter $C$ for ``standard'' SVM (that is, without
individual instance weights) and the individual instance weights $C_i$
for weighted SVM---is important in order to achieve good
generalization performance. For standard SVM, various hyperparameter
selection methods have been proposed and evaluated: ordinary $K$-fold
cross-validation ($K$-fold CV) and leave-one-out procedure; genetic
based methods \citep{huang2006ga}; particle optimization methods
\citep{lin2008particle}; and generalized approximate cross-validation
\citep{wahba1999bias}.

In addition to the previous, upper bounds or estimators for the
leave-one-out error aim to provide computationally cheaper
alternatives of the actual leave-one-out procedure
\citep{jaakkola1999probabilistic,opper2000gaussian,Joachims2000estimating}.
\cite{vapnik2000bounds} introduced the concept of the \emph{span of a
  support vector} which is a geometrical object defined using the
support vectors of a trained SVM model. Based on the span, they proved
the \emph{span bound}, an upper bound of the leave-one-out error, for
standard soft- and hard- margin SVM algorithms. Additionally, they
proved the \emph{span-rule}, a method that estimates the exact value
of the leave-one-out error, and demonstrated its use for effective
hyperparameter selection.

Although there are various leave-one-out error bounds and
leave-one-out error prediction methods for standard soft- and hard-
margin SVM, the existence of leave-one-out error estimators suitable
for weighted SVM algorithm is fairly limited
\citep{yang2015confirming}. The related work on leave-one-out error
estimators and bounds for standard and weighted SVM is given in detail
in Section \ref{sec:related_work}.

\subsection{Contribution of this work}
This work extends the span error bound theory to weighted SVM
algorithm. The new span bound and span-rule contribute both in
theoretical understanding and in practical hyperparameter selection
applications for this widely-used SVM variant.

More specifically, the key theoretical contributions of this work are:
\begin{itemize}
\item We prove in Lemma 1 the necessary and sufficient condition for
  the existence of the span of a support vector in a weighted SVM
  model. The importance of this lemma is that allows the extension of
  the span error bound theory to weighted SVM.
\item We give in Theorem 1 the extension of the span bound (upper
  bound of the test error) to weighted SVM. The span bound takes into
  account both the support vectors that satisfy Lemma 1 and those that
  do not.
\item We prove in Theorem 2 that, under the assumption that the
  support vectors do not change during the leave-one-out procedure,
  the condition of Lemma 1 is satisfied for \emph{all} support vectors
  of a weighted SVM model. Also, Theorem 2 confirms the equality of
  \cite{vapnik2000bounds} that is useful for identifying a support
  vector as a leave-one-out error.
\item We confirm in Corollary 1 the extension of the span-rule of
  \cite{vapnik2000bounds} to weighted SVM algorithm.
\end{itemize}

The experimental contributions verify the practical importance of the
extended theory:
\begin{itemize}

\item We experimentally verify that the existence condition for the
  span of a support vector, given by Lemma 1, is satisfied almost
  always. This observation justifies the extension of the span error
  bound theory to weighted SVM algorithm and its usage in practical
  applications.

\item We experimentally evaluate the effectiveness and efficiency of
  the span bound and the span-rule in comparison with other hyperparameter
  selection methods applicable to weighted SVM: the $K$-fold CV and
  the $\xialpha$ bound of \cite{yang2015confirming}.

\item
  We compare the hyperparameter selection methods in two problems: (a)
  hyperparameter selection for class weights (that is, $C_i = C^+$ if $y_i
  = +1$ and $C_i = C^-$ if $y_i = -1$); and (b) hyperparameter selection to
  identify the optimal mapping from individual importance scores of
  the training instances, $q_i$, to individual instance weights,
  $C_i$.

\item Effectiveness and efficiency experiments conducted on 14
  standard benchmark data sets and data sets with instance importance
  scores demonstrate that: (a) the span-rule for weighted SVM is the
  best hyperparameter selection method in both examined problems; (b) the
  span-rule is the best predictor for the value of the actual testing
  error in the mean-square-error sense; and (c) the span-rule
  computation time is not only compelling but, in certain cases, it
  can be significantly faster than $K$-fold CV.

\end{itemize}

\subsection{Organization}
The rest of the paper is organized as follows. Section
\ref{sec:preliminaries} briefly presents the preliminaries on weighted
SVM formulation and the theorem for the unbiasedness of the
leave-one-out error estimator. Then, Section \ref{sec:related_work}
presents an overview of the related work on leave-one-out error bounds
for SVM-based algorithms. Section \ref{sec:span_theory} presents the
span bound and span-rule for weighted SVM. Next, Section
\ref{sec:weight_optimization} illustrates the proposed hyperparameter
selection approach using the span-rule.  Finally, Section
\ref{sec:experiments} presents the experimental analysis and Section
\ref{sec:conclusions} summarizes the findings and concludes the paper.

\section{Preliminaries}
\label{sec:preliminaries}
This section briefly presents the preliminaries on weighted SVM
(Section \ref{subsec:weighted_svm_prelim}) and the theorem for the
unbiasedness of the leave-one-out error estimator (Section
\ref{subsec:loo_prelim}).

\subsection{The weighted SVM algorithm}
\label{subsec:weighted_svm_prelim}

Let $\T_l$ be a training set of $l$ instances, $\T_l = \{(\x_1,y_1),
\dots,$ $ (\x_l,y_l)\}$, where $\x_i \in \mathbb{R}^d$ are the
$d$-dimensional feature vectors and $y_i \in \{-1,+1\}$ are the class
labels of the instances.

We consider the weighted SVM algorithm where a realization of the
soft-margin separating hyperplane with threshold, $\w \cdot \x + b =
0$, is computed from \citep{lin2002fuzzy}:
\begin{align}
  \min ~ \frac{1}{2}\w \cdot \w &+ \sum_{i=1}^l C_i \xi_i  \label{eq:fsvm_obj} \\
  s.t. ~~ y_{i} (\w \cdot \x_i + b) &\geq 1 - \xi_i,
  ~~~i=1,\dots,l \label{eq:fsvm_constrain1} \\
  \xi_i &\geq 0, ~~~i=1,\dots,l \label{eq:fsvm_constrain2}
\end{align}

The slack variables $\xi_i$ allow the training instances to violate
the margin; a training instance violates the margin when $\xi_i > 0$
and is misclassified when $\xi_i \geq 1$. 
The instance weights $C_i$ regularize the training error penalties in
the objective function. A larger value for a $C_i$ makes more likely
$\xi_i = 0$ to occur.

The difference between the formulations of standard SVM and weighted
SVM is that the latter uses the individual instance weights $C_i$ to
regularize the effect of the training errors with a different
intensity for each instance, whereas, for standard SVM it is $C_i = C$
for all training instances.

Equations \eqref{eq:fsvm_obj}, \eqref{eq:fsvm_constrain1} and
\eqref{eq:fsvm_constrain2} define the primal problem of weighted
SVM. The Lagrangian of the primal problem is:
\begin{equation}
\label{eq:fsvm_primal}
\Lagr_P = \frac{1}{2}\w \cdot \w + \sumL C_i \xi_i -
\sumL \a_i [ y_i ( \w \cdot \x_i + b ) - 1 + \xi_i ] -
\sumL \mu_i \xi_i 
\end{equation}
where $\a_i$ and $\mu_i$ are the Lagrange multipliers from
inequalities \eqref{eq:fsvm_constrain1} and \eqref{eq:fsvm_constrain2}
respectively.

Taking the first order derivatives of $\Lagr_P$ with respect to $\w$,
$b$ and $\xi_i$, setting them to zero and substituting back to
\eqref{eq:fsvm_primal}, leads to the dual problem for weighted SVM:
\begin{align}
\max ~ W(\ba) \equiv \sumL \a_i &- 
\frac{1}{2} \sumLij \a_i \a_j y_i y_j \x_i \cdot \x_j \label{eq:fsvm_dual_obj} \\
  s.t. ~~  0 \leq \a_i &\leq C_i  \label{eq:fsvm_dual_con1} \\
   \sumL \a_i y_i &= 0  \label{eq:fsvm_dual_con2}
\end{align}

The Karush-Kuhn-Tucker (KKT) conditions for weighted SVM are:
\begin{align}
\frac{\partial \Lagr_P}{ \partial \w} = \w - \sumL \a_i y_i \x_i &= 0 \label{eq:KKT_deriv1}\\
\frac{\partial \Lagr_P}{ \partial b} = -\sumL \a_i y_i &= 0 \\
\frac{\partial \Lagr_P}{ \partial \xi_i} = C_i - \a_i - \mu_i &= 0 \label{eq:KKT_deriv3}\\
y_{i} (\w \cdot \x_i + b) - 1 + \xi_i &\geq 0 \\ 
\xi_i &\geq 0 \\ 
\a_i &\geq 0 \\
\mu_i &\geq 0 \\
\a_i \{ y_{i} (\w \cdot \x_i + b) - 1 + \xi_i \} &= 0 \label{eq:KKT_complementary1} \\
\mu_i \xi_i &= 0 \label{eq:KKT_complementary2}
\end{align}
Equations \eqref{eq:KKT_complementary1} and
\eqref{eq:KKT_complementary2} represent the KKT complementary
slackness conditions.

For a support vector with $0 < \a_i <C_i$, equations
\eqref{eq:KKT_deriv3} and \eqref{eq:KKT_complementary2} result to
$\xi_i = 0$. Also, from the same equations follows that when a support
vector violates the margin ($\xi_i > 0$) then $\a_i = C_i$.

We will use the notation of \cite{vapnik2000bounds} to discriminate
the categories of the support vectors. The instances of the training
set will be sorted so that the first $n^*$ are the \emph{in-bound}
support vectors ($0 < \a_i <C_i$) followed by the $m = n-n^*$
\emph{bounded} support vectors ($\a_i = C_i$). We consider only stable
solutions where $n^*>0$; an assumption that generally holds true
\citep{crisp2000uniqueness}.

There is no difference between weighted SVM and standard SVM for
predicting the label of a new instance. That is, a new observation
$\x$ is classified according to the sign of
\begin{equation}
\label{eq:fsvm_decision_function}
f(\x) = \w \cdot \x + b = \sumL \a_i y_i \x_i \cdot \x + b
\end{equation}

SVMs can be extended to produce non-linear solutions via the kernel
trick, $K(\x, \x') = \phi(\x) \cdot \phi(\x')$, where
$\phi:\mathbb{R}^d \mapsto \mathcal{H}$ produces a mapping of the
initial feature vectors to a higher dimensional space,
$\mathcal{H}$. Section \ref{sec:span_theory} will assume the linear
kernel, $K(\x, \x') = \x \cdot \x'$; however, the presented theory can
be directly extended to non-linear SVMs.

\subsection{The leave-one-out error theorem}
\label{subsec:loo_prelim}
In the problem of learning from examples, the learning machine is
supplied with a training set of $l$ instances, $\T_l = \{(\x_1,y_1),
\dots,$ $ (\x_l,y_l)\}$, drawn i.i.d. from an unknown underlying joint
distribution $F(\x,y)=F(\x)F(y|\x)$.

Let us denote with $R(f_{T_l})$ the test error rate of a machine,
$f$, trained on a sample $T_l$. One can use the $l$ instances of
$\T_l$ to calculate an estimate of $R(f_{T_l})$ through a
leave-one-out procedure.
In the leave-one-out procedure, one instance $(\x_i,y_i)$ is omitted
from the training set, a decision function is trained using the
remaining $l-1$ instances and the error on the left-out instance is
calculated. The procedure is repeated for each instance of the
training set and the total number of errors is calculated. 

Let $\Lagr(\x_1,y_1, \dots, \x_l,y_l)$ be the total number of errors
from the leave-out-procedure on $T_l$. The \emph{leave-one-out error}
$\LOOError(\T_l)$ is given by
\begin{equation}
        \label{eq:classical_Rloo}
        \LOOError(\T_l) = \frac{1}{l} \Lagr(\x_1,y_1, \dots, \x_l,y_l)
\end{equation}

The leave-one-out error is an almost unbiased estimator of the test
error \citep{vapnik1998statistical,luntz1969techni}, that is:
\begin{equation}
  \label{eq:loo_theorem}
E_{\T_l}[\LOOError(\T_l)] = E_{\T_{l-1}}[R(f_{T_{l-1}})]
\end{equation}
where the expectation $E_{\T_l}[\LOOError(\T_l)]$ is taken over the
ensemble of training sets of size $l$ and
$E_{\T_{l-1}}[R(f_{T_{l-1}})]$ is the expectation of the test error
for the learning machine trained on $l-1$ instances. The ``almost''
refers to the fact that the test error estimated by leave-one-out
procedure is for the learning machine trained on sets of size $l-1$
instead of $l$.

The leave-one-out theorem (eq. \ref{eq:loo_theorem}) is the basis of
the examined error bounds and error estimators presented next.

\section{Related Work}
\label{sec:related_work}

This section gives an overview of the leave-one-out error bounds and
leave-one-out error estimators for SVM-based algorithms.

Although the leave-one-out error is an almost unbiased estimator of
the test error, its actual calculation is generally impractical due to
high computational cost. To this end, a number of approximations or
upper bounds have been proposed that attempt to estimate the
leave-one-out error using only a single SVM model that is trained with
all available instances of the training set $\T_l$. Using these
methods we do not need to train additional SVM models; whereas, this
is required by other hyperparameter selection methods (such as, the
$K$-fold CV).

A training instance that is not a support vector is correctly
classified by the SVM model. Also, since the SVM solution is unique,
removing from $\T_l$ a training instance that is not support vector
leaves the hyperplane unchanged. Thus, the number of support vectors,
$n$, is an upper bound of the leave-one-out errors
\citep{vapnik1998statistical},
\begin{equation}
\label{eq:sv_count_bound}
\LOOError(\T_l) \leq \frac{n}{l}
\end{equation}

For the separable case and hard-margin SVM without threshold (that is,
separating hyperplanes in the form: $\w \cdot \x = 0$),
\cite{vapnik1998statistical} proved the radius-margin bound
\begin{equation*}
\LOOError(\T_l) \leq \frac{1}{4l}D^2 ||\w||^2
\end{equation*}
where $D$ is the diameter of the smallest sphere containing the
training set. 

\cite{chapelle2002choosing} calculated the gradient of the
radius-margin bound and showed that it can be used for SVM parameter
tuning and feature selection via gradient descent methods over the
parameter set.

A modified version of radius-margin bound for soft-margin SVM was
studied by \cite{duan2003evaluation} and showed adequate performance
for the problem of hyperparameter selection. A version of radius-margin
bound for $L_{2}$ soft-margin SVM was proposed by
\cite{keerthi2002efficient}.

For hard-margin SVM, \cite{jaakkola1999probabilistic} used
probabilistic regression models to prove an upper bound of the
leave-one-out error
\begin{equation*}
\LOOError(\T_l) \leq \frac{1}{l} \sum\limits_{p=1}^l
\theta\left(y_p[f(\x_p) - \a_pK(\x_p,\x_p)]\right)
\end{equation*}
where $\a_p$ are the Lagrange multipliers of the support vectors,
$\theta$ is the step function ($\theta(t) = 1 \text{ when } t\geq 0
\text{ and } \theta(t) = 0 \text{ otherwise}$), $K(\x,\x')$ is the
kernel function and $f(\x)$ is the decision function.

\cite{opper2000gaussian} proposed an approximation of the
leave-one-out error based on methods from statistical physics. Under
the assumption that the set of in-bound and the set of bounded support
vectors do not change during leave-one-out procedure, an estimate of
the leave-one-out error is given by
\begin{equation*}
t_l = \frac{1}{l} \sum\limits_{p=1}^l \theta\left(\frac{a_p}{(\mathbf{K}_{SV}^{-1})_{pp}} -1\right)
\end{equation*}
where $\mathbf{K}_{SV}$ is the Gram matrix of the support vectors.

Similar to the Jaakkola-Haussler bound, the $\xialpha$ \emph{bound}
for the soft-margin SVM algorithm with threshold was proposed by
\cite{Joachims2000estimating}. This bound is calculated using the
Lagrange multipliers of the support vectors, $\a_p$, and the slack
variables $\xi_p$. Although $\xialpha$ bound is fairly conservative it
was effectively used for hyperparameter selection with SVM-based text
classifiers.

\cite{yang2015confirming} showed that the $\xialpha$ bound can be
applied to weighted SVM as well. For the weighted SVM formulation of
Section \ref{subsec:weighted_svm_prelim}, the $\xialpha$ error bound
is given by
\begin{equation}
\label{eq:xialpha_rule_fsvm}
\LOOError(\T_l) \leq \frac{1}{l} \sum\limits_{p=1}^l \theta \left( 2\a_p R_{\Delta}^2 + \xi_p - 1 \right)
\end{equation}
where $R_{\Delta}^2$ satisfies $c\leq K(\x_i, \x_j) \leq c +
R_{\Delta}^2$ for all $\x_i, \x_j$ and a constant $c$.

Note that the $\xialpha$ bound is the only leave-one-out error bound
in the literature that can be applied with weighted SVM. To this end,
the bound of \eqref{eq:xialpha_rule_fsvm} is experimentally evaluated
in this work for comparison purposes.

\cite{vapnik2000bounds,chapelle2000modelselection} introduced the
concept of the \emph{span of a support vector}---a geometrical object
defined using the remaining in-bound support vectors of the SVM
solution when a support vector is left out---and used it to derive the
span bound and the span-rule. The concept of the span and its
calculation will be described in detail in Section
\ref{sec:span_theory}.

If $S_p$ is the distance of the $p$-th left-out support vector,
$\x_p$, from the corresponding geometrical object then the span bound
is given by
\begin{equation}
  \LOOError(\T_l) \leq \frac{S \sum_{p=1}^{n^*}[\max(D,1/\sqrt{C})\a_p^0] + m}{l}
\end{equation}
where $S = \max_{\substack {p = 1, \dots, n^*} } S_p$ and $n^*$ is the
number of in-bound support vectors.

Also, under the assumption that the sets of in-bound and bounded
support vectors do not change in leave-one-out procedure, the
span-rule calculated by
\begin{equation}
\label{eq:related_work_span_rule}
t_l = \frac{1}{l} \sum\limits_{p=1}^l \theta \left(  \a_p S_p^2 - y_p f(\x_p) \right)
\end{equation}
gives the exact value of the leave-one-out error; that is,
$\LOOError(\T_l) = t_l$.

The intuition behind span bound and span-rule is that smaller
distance, $S_p$, and smaller Lagrange multiplier, $\a_p$, indicate
that the support vector is less likely to be a leave-one-out error.

Additionally, it was shown that the value of $S_p$ is smaller than the
diameter of the sphere that encloses the in-bound support vectors and,
for this reason, the span-rule is considered to give accurate error
estimates
\citep{vapnik2000bounds,duan2003evaluation,scholkopf2002learning}. Regarding
the efficiency, it was shown that the use of span-rule can be
significantly faster than executing an actual leave-one-out procedure.

In this work, we prove the necessary and sufficient condition for the
existence of the span of a support vector in a weighted SVM solution
(in Lemma 1) and we show that the span bound and the span-rule can be
defined for weighted SVM as well (in Theorems 1 and 2 respectively).
Extensive experimental analysis shows that the span-rule is an
effective and efficient hyperparameter selection method for the
weighted SVM algorithm.

\section{Span Bound and Span-Rule for Weighted SVM}
\label{sec:span_theory}
This section presents the extension of the span error bound theory to
weighted SVM. First, in Section \ref{subsec:span_existence} we define
the \emph{span of a support vector} for weighted SVM. Most
importantly, we provide in Lemma 1 the non-trivial proof for the
necessary and sufficient condition for the existence of the span of a
support vector in a weighted SVM model. Then, in Section
\ref{subsec:span_bound} we extend the span bound to weighted SVM.

Next, in Section \ref{subsec:span_rule} we give in Corollary 1 the
span-rule for weighted SVM. Notably, we prove in Theorem 2 that, under
the assumption that the support vectors do not change during
leave-one-out procedure, the value of the span can be calculated for
\emph{any} support vector; that is, the existence condition of Lemma 1
is always satisfied. Thus, the span-rule for weighted SVM is applied
using all the support vectors, similar to the span-rule for standard
SVM.

\subsection{Span of a support vector in weighted SVM}
\label{subsec:span_existence}
Let us consider that the instances of the training set $\T_l$ are
ordered so that the first $n$ ones are the support vectors,
$(\x_1,y_1), \dots, (\x_n,y_n)$. Also, let $\aV^0 = (\a_1^0, \dots,
\a_n^0, 0, \dots, 0)$ be the vector of the Lagrange multipliers for
the weighted SVM solution trained on the initial training set (that
is, using all instances of $\T_l$). 

We consider that the supports vectors are also ordered, so that the
first $n^*$ are the in-bound support vectors ($0 < \a_i^0 < C_i, ~ i =
1,\dots, n^*$) and the remaining $m = n-n^*$ are the bounded support
vectors ($\a_i^0 = C_i, ~ i = n^*+1, \dots, n $).

The in-bound support vectors are used for defining the set $\Lp$. For
an in-bound support vector $\x_p$, we define the set $\Lp$ using the
remaining in-bound support vectors as:
\begin{align}
\label{eq:SVMLambdaSet}
\Lambda_p = \Bigg \{ \sumUBNoP \ll_i \x_i : \sumUBNoP \ll_i = 1,
  \text{and } 
   0 \leq \a^0_i + y_i y_p \a^0_p \ll_i \leq C_i ~~
  \forall i=1,\dots,n^*, ~ i \neq p
 \Bigg \}
\end{align}
where $\ll_i$ are real-valued variables. 

The set $\Lp$ is the basis of the span error bound theory for weighted
SVM. We can acquire the corresponding set for standard SVM by setting
$C_i = C$.

For the standard SVM, it was shown by \cite{vapnik2000bounds} that the
set $\Lp$ is non-empty for all support vectors. However, as we will
prove next, this is not true for the weighted SVM algorithm. Thus, in
order to extend the span error bound theory to weighted SVM we need to
examine when the set $\Lp$ can be defined (that is, the conditions
assuring that $\Lp \neq \emptyset$).

The following lemma provides the necessary and sufficient condition
for the existence of $\Lp$ which is needed to extend the span error
bound theory to weighted SVM.

\vskip 0.1in
\noindent
{\bf Lemma 1} {\it The set $\Lp$ for a fixed in-bound support vector
$\x_p$ of a weighted SVM solution is non-empty if and only if }
\begin{equation}
  \label{eq:existence_condition}
	\sumUBNoPpos C_i + y_p \sumB y_i C_i \geq 0 
\end{equation} \hfill $\BlackBox$

Proof is in Appendix A.1.
\vskip 0.1in

The importance of Lemma 1 lies on the fact that without it we would
not be able to extend the span error bound theory to weighted SVM algorithm.

\vskip 0.05in

For weighted SVM we define the span $S_p$ of a support vector $\x_p$
as its distance from the set $\Lp$ when the set is non-empty,
\begin{equation}
\label{eq:SVMSpan}
S_p^2 = d^2(\x_p, \Lp) = 
\min_{\x \in \Lp} (\x - \x_p)^2, ~~ \Lp \neq \emptyset
\end{equation}

The support vector $\x_p$ in \eqref{eq:SVMSpan} can be in-bound or
bounded.

If the necessary and sufficient condition of Lemma 1 does not hold for
an in-bound support vector $\x_p$ then we consider that the value of
$S_p$ cannot be calculated for this support vector. Nevertheless,
experimental analysis of weighted SVM models on various data sets
shows that the occurrence of $\Lp = \emptyset$ is a very rare
exception (see Section \ref{subsec:existence_Lp}).

The property of standard SVM for the span
\begin{align}
\label{eq:SpInequality}
S_p \leq D_{SV}
\end{align}
holds for weighted SVM as well (see remark of proof, Appendix
A.1). $D_{SV}$ is the diameter of the smallest sphere enclosing the
in-bound support vectors.

Using \eqref{eq:SVMLambdaSet}, \eqref{eq:SVMSpan}, setting $\ll_p = -1$
and $\ll_i = 0 \text{ for } i>n^* \text{, } i \neq p~$ we can calculate
the value of $S_p^2$ for any support vector from the following:
\begin{align}
  \label{eq:Sp_calc_1}
   & S_p^2 \equiv \min_{\ll_i} \left( \sumN \ll_i  \x_i \right)^2 \\
  \text{ s.t. }  & \ll_p = -1 \\
   & \sumN \ll_i = 0 \label{eq:linear_con} \\
   & 0 \leq \a^0_i + y_i y_p  \a^0_p \ll_i \leq C_i, ~~~ \forall i \neq p \label{eq:box_con} \\
  \label{eq:Sp_calc_4}
   & \ll_i  = 0, ~~~ \forall i >n^* \text{ and } i \neq p
\end{align}

As we will see later, under the assumption that the support vectors do
not change during leave-one-out procedure, the box constrains of
\eqref{eq:box_con} may be dropped.

Finally, we define the $S$-span for weighted SVM as the maximum value
of $S_p$ among the in-bound support vectors with $\Lp \neq \emptyset$,
\begin{equation}
\label{eq:SSpan}
S = \max_{\substack {p = 1, \dots, n^* \\ \Lp \neq \emptyset} } S_p
\end{equation}

\subsection{Span bound for weighted SVM}
\label{subsec:span_bound}
In this section, we present the extension of the span bound to
weighted SVM. We begin with the following lemma:

\vskip 0.1in
\noindent
{\bf Lemma 2} {\it If in the leave-one-out procedure an in-bound
  support vector} ($0 < \a_p^0 < C_p, ~~ p = 1, \dots, n^*$) {\it with }
$\Lp \neq \emptyset$ {\it is misclassified, then the inequality
\[
     \a_p^0 S_p \max \left(D, \frac{1}{\sqrt{C_p}} \right) \geq 1
\]
holds true. $D$ is the diameter of the smallest sphere enclosing the
 training set.}
\hfill $\BlackBox$

This proof is omitted due to similarity with the corresponding proof
for standard SVM in the non-separable case \citep[see Lemma 2.3
  of][]{vapnik2000bounds}. The only differences in executing the proof
of this lemma originate from the individual instance weights $C_i$.
\vskip 0.1in

Let us denote with $k$ the number of in-bound support vectors where
the condition of Lemma 1 is not satisfied and, thus, for these support
vectors it is $\Lp = \emptyset$. Also, suppose that the in-bound
support vectors are further ordered so that the first $n^*-k$ in-bound
support vectors are the ones with $\Lp \neq \emptyset$.

Using Lemma 1 and Lemma 2 we can prove the span bound theorem for the
expectation of the test error for weighted SVM.

\vskip 0.1in 
\noindent
{\bf Theorem 1} {\it The expectation of the test error
  $E_{\T_{l-1}}[R(\a^{(l-1)})]$ for a weighted SVM model trained on
  $l-1$ instances has the bound
\[
E_{\T_{l-1}}[R(\a^{(l-1)})] \leq E_{\T_l} \left[ 
  \frac{S  \sum_{p=1}^{n^*-k}[\max(D,1/\sqrt{C_p})\a_p^0] + k + m}{l} \right]
\]
where $n^*-k$ is the number of in-bound support vectors that satisfy
the inequality of Lemma 1 } ({\it that is, the support vectors with }
$\Lp \neq \emptyset$), {\it $k$ is the number of the in-bound support
  vectors where the inequality of Lemma 1 is not satisfied } ({\it
  that is, the support vectors with } $\Lp = \emptyset$), {\it and
  $m=n-n^*$ is the number of the bounded support vectors. The values,
  $S,$ of S-span, the diameter, $D,$ of the sphere containing the
  training set and the Lagrange multipliers, $\a_p^0$, of the
  weighted SVM model are considered for training set of size $l$.}
\hfill $\BlackBox$

Proof is in Appendix A.2.
\vskip 0.1in 

Theorem 1 can be viewed as an extension of the original span
bound. Indeed, since for standard SVM it is always $k=0$, we can
acquire the span bound for the non-separable case of standard SVM.

Because the $k$ in-bound support vectors with $\Lp = \emptyset$ are
counted as leave-one-out errors, the error bound of Theorem 1 appears
to be less tight when applied to weighted SVM than standard SVM.
More specifically, if the equation of Lemma 1 is not satisfied for any
of the in-bound support vectors (that is, if $k=n^*$), then the error
bound becomes
\begin{equation}
\label{eq:upper_span_value}
E_{\T_{l-1}}[R(\a^{(l-1)})] \leq E_{\T_{l-1}}\left[ \frac{n^* +
m}{l} \right] = E_{\T_{l-1}} \left[ \frac{n}{l} \right]
\end{equation}

Thus, in this case, the span bound degenerates to the
support-vector-count error bound given by
\eqref{eq:sv_count_bound}. Nevertheless, the experimental results show
that the occurrence of $\Lp = \emptyset$ is very rare and the error
bound of Theorem 1 is virtually unaffected by these support vectors.

\subsection{Span-rule for weighted SVM}
\label{subsec:span_rule}
The bound defined by Theorem 1 is always available, however, by posing
realistic assumptions we can reach an exact estimate of the test error
instead of an error bound.

The span-rule, presented next, gives the value of the leave-one-out
error under the assumption that the support vectors do not change
during the leave-one-out procedure \citep{vapnik2000bounds}. The same
assumption was made in deriving the bounds of
\cite{opper2000gaussian}.

We begin by proving the following theorem for weighted SVM:

\vskip 0.1in 

\noindent
{\bf Theorem 2 } {\it Under the assumption that the sets of the
  in-bound and bounded support vectors remain the same during the
  leave-one-out-procedure, then for any support vector $\x_p$ the
  inequality of Lemma 1 is always satisfied and $\Lp \neq
  \emptyset$. Furthermore, for any support vector, the following
  equality holds true:
\begin{equation}
y_p (f^0(\x_p) - f^p(\x_p)) = \a_p^0 S_p^2 \label{eq:theorem4}
\end{equation}
where $f^0$ and $f^p$ are the decision functions for the weighted SVM
trained on the whole training set and the training set after the
removal of $\x_p$ respectively.} \hfill $\BlackBox$

Proof is in Appendix A.3.

\vskip 0.1in 

Theorem 2 denotes that, under the assumption that the support vectors
do not change during leave-one-out procedure, the inequality of Lemma
1 is always satisfied. This is an important finding since the
theorem's assumption is generally true for most of the support
vectors. Theorem 2 also aligns with the experimental observation that
the inequality of Lemma 1 is satisfied almost always (see Section
\ref{subsec:existence_Lp}).

Similarly to the corresponding theorem for standard SVM, other gains
of Theorem 2 compared to Theorem 1 are: (a) the inequality becomes an
equality, (b) the value $D$ diminishes and instead the smaller value
of $S_p$ appears in the equation; (c) the theorem can be applied for
all support vectors; and (d) the box constrains of
\eqref{eq:SVMLambdaSet} are trivially satisfied and can be dropped
from the calculation of $S_p$ (as shown in first part of theorem's
proof).

\vskip 0.05in 

From the equality \eqref{eq:theorem4} we deduce that a leave-one-out
error (that is, $y_p f^p(\x_p) \leq 0$) will occur for support vector
$\x_p$ when
\begin{equation}
                \label{eq:iff}
 \a_p^0 S_p^2 \geq y_p f^0(\x_p), ~~ p = 1,\dots, n
\end{equation}

Using the leave-one-out theorem (eq. \ref{eq:loo_theorem}) and taking
into account inequality \eqref{eq:iff} we get the span-rule for
weighted SVM.

\vskip 0.1in 
\noindent 
{\bf Corollary 1 -- Span-rule} {\it Under the assumption of Theorem 2,
  the leave-one-out error, $\LOOError$, is equal to
\begin{align}
\label{eq:span_rule}
t_l \equiv \frac{1}{l} \sum\limits_{p=1}^n \theta
\left(\a_p^0 S_p^2 - y_p f^0(\x_p) \right)
\end{align}
}
where $\theta$ is the step function; $\theta(t) = 1 \text{ when }
t\geq0 \text{ and } \theta(t) = 0 \text{ otherwise}$.
\vskip 0.1in 

The span-rule of Corollary 1 is applied using \emph{all} support
vectors since Theorem 2 guarantees the existence of the span, $S_p$,
for all support vectors under the theorem's assumption.

Note that, contrary to the actual leave-one-out procedure, the
computation of $t_l$ requires training only one weighted SVM model;
however, we do need to compute $n$ different values of $S_p$. Sections
\ref{subsec:computational_comp} and \ref{subsec:when_faster} examine
when this is computationally attractive.

\vskip 0.05in 

We can apply the span-rule in the following practical problems:
\begin{itemize}
\item Given a trained weighted SVM model, we can estimate the value of
  the testing error.

\item Most importantly, we can use the span-rule as a typical
  hyperparameter selection tool. That is, we can train weighted SVM models
  with different parameterizations (for example, different weights
  $C_i$) and select the model with the best generalization
  performance. Next section sketches the proposed approach for
  weighted SVM hyperparameter selection using the span-rule.

\end{itemize}

\section{Hyperparameter selection using the span-rule}
\label{sec:weight_optimization}
The goal of hyperparameter selection (or model selection) is to
discover the training hyperparameters that yield the model with the
best generalization performance; that is, the model that will be able
to classify unknown instances with the smallest classification error.

\begin{figure}[!ht]
\centering
\includegraphics[width=0.95\textwidth]{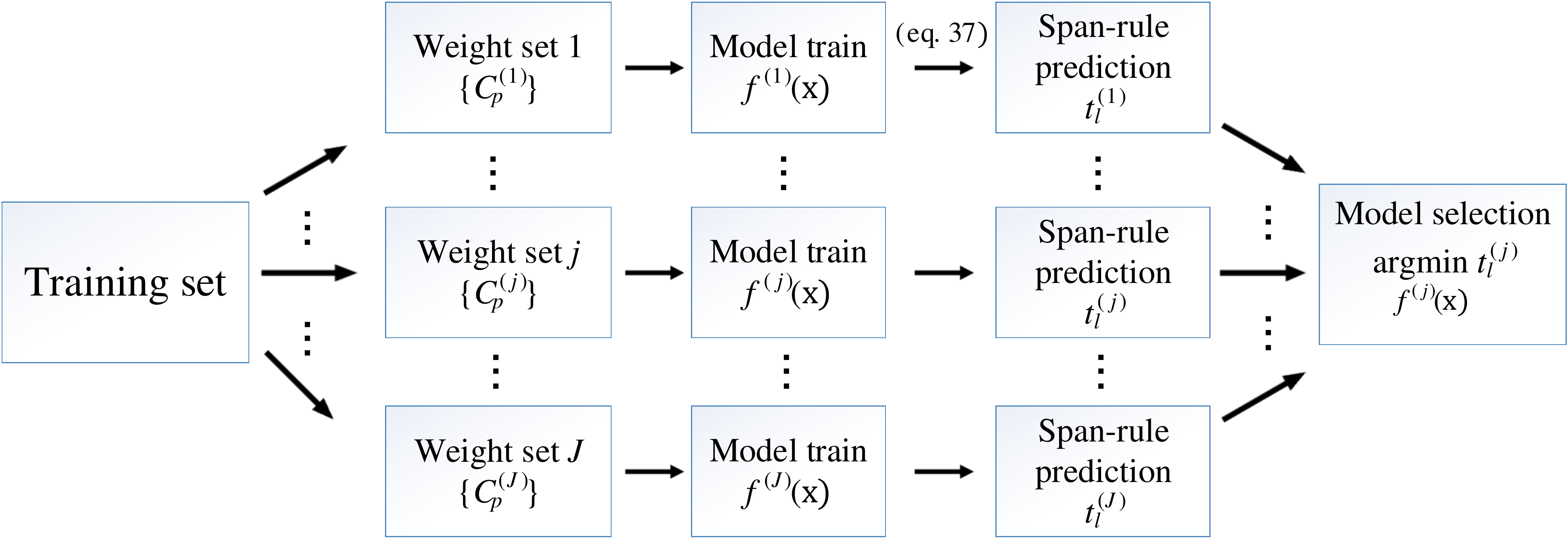}
\caption{Sketch of the proposed approach for hyperparameter selection. Each
  one of the $J$ branches produces a final, weighted SVM model with an
  estimation of its test error.}
\label{fig:optimization_sketch}
\end{figure}

This work enables a new method to perform hyperparameter selection for
weighted SVM that was not previously possible. A sketch of the
proposed method for instance weights selection is illustrated in
Figure \ref{fig:optimization_sketch}.

\vskip 0.1in

At $j$-th iteration, we execute the following three steps:

\begin{enumerate}

\item \emph{Generation of training hyperparameters}. A set of instance
  weights, $\{C^{(j)}_p, ~ p =1,\dots,l\}$, is generated. Also, other
  training hyperparameters may be set, such as the kernel-specific
  parameters for non-linear SVMs.

\item \emph{Weighted SVM model training}. A weighted SVM model with
  decision function $f^{(j)}(\x)$ is trained using \emph{all}
  available training instances and the weights generated in the
  previous step.

\item \emph{Test error prediction using span-rule}. Using the
  span-rule, the error prediction, $t_l^{(j)}$, is calculated for the
  $j$-th candidate weighted SVM model.

\end{enumerate}

We repeat these steps for a total of $J$ parameterizations. In the
end, $J$ candidate weighted SVM models with decision functions
$f^{(j)}(\x)$ are available and each model is accompanied with a
prediction for the testing error of the model. This prediction
reflects an assessment of its generalization performance.

The final step is to select the model with the best performance
according to
\begin{equation}
\argmin \limits_{\substack{f^{(j)}(\x)}} t^{(j)}_l, ~~~ j = 1, \dots, J
\end{equation}

Although other hyperparameter selection methods can be used for test
error estimation in the place of span-rule in step 3, the experiments
demonstrate that the span-rule is the most effective and, in certain
cases, the most efficient method for the weighted SVM algorithm.


\section{Experiments}
\label{sec:experiments}
This section presents the experimental analysis and is organized as
follows:
\begin{itemize}
\item Section \ref{subsec:datasets} gives an overview of the data
  sets. Two types of data sets are used: (a) standard benchmark data
  sets; (b) data sets where a problem-specific, importance score $q_i$
  is known for each training instance.

\item Section \ref{subsec:cplusminus} ({\it Experiment \#1}) presents
  the experimental evaluation of the span bound of Theorem 1 and the
  span-rule of Corollary 1 on the task of selecting optimal class
  weights $C^+/C^-$. The standard benchmark data sets are used. The
  methods are compared to other hyperparameter selection methods
  applicable to weighted SVM: the $K$-fold CV (with $K=5$, which is a
  fairly common choice) and the $\xialpha$ bound for weighted SVM.

\item Section \ref{subsec:weight_selection} ({\it Experiment \#2})
  presents the evaluation of the four examined hyperparameter
  selection methods on the task of selecting optimal, individual
  instance weights $C_i$ using the importance enconded in the scores
  $q_i$. The data sets with problem-specific importance scores
  assigned for each training instance are used.

\item Section \ref{subsec:predicting_curves} ({\it Experiment \#3})
  presents the evaluation of the span-rule, $K$-fold CV and the
  $\xialpha$ bound on the task of predicting the actual value of the
  test error. Both types of data sets are used for this evaluation.

\item Section \ref{subsec:existence_Lp} ({\it Experiment \#4})
  presents an experimental investigation on the satisfaction of the
  inequality of Lemma 1 that guarantees the non-emptiness of the set
  $\Lp$ and, consequently, the existence of the span, $S_p$. Both
  types of data sets are used.

\item Section \ref{subsec:computational_comp} ({\it Experiment \#5})
  presents the efficiency evaluation of the two most prominent
  methods: the span-rule and the $K$-fold CV. Both types of data sets
  are used.

\item Section \ref{subsec:when_faster} ({\it Experiment \#6}) aims to
  answer when it is faster to use span-rule than $K$-fold CV through
  experiments on synthetic data with varying training set sizes and
  feature dimensionality.

\item Section \ref{subsec:exp_summary} summarizes the most
  important experimental findings.
  
\end{itemize}

\paragraph{Implementation details.}%
We use LibSVM \citep{Chang2011libsvm} for training the weighted SVM
models and executing the $K$-fold CV. The default cache size of
$100$MB is used. The MOSEK optimizer \citep{mosek} is used for
calculating the value of the span $S_p$ by solving
\eqref{eq:Sp_calc_1}-\eqref{eq:Sp_calc_4}. Under the assumption of
Theorem 2, we can drop the box constrains (eq. \ref{eq:box_con}) for
the calculation of $S_p$ for span-rule (as shown in first part of
theorem's proof, Appendix A.3).  Finally, the implementation of
$\xialpha$ bound of \eqref{eq:xialpha_rule_fsvm} is fairly simple as
it can be calculated directly from the trained SVM models.

\subsection{Data sets}
\label{subsec:datasets}
Table \ref{table:datasets} displays an overview of the data
sets. Following the suggestion of \cite{duan2003evaluation}, we use
large testing sets in order to achieve accurate assessments of the
generalization performance.  In this section, we will refer to the
error rate on the testing sets as ``test error''. Two types of data
sets are used: standard benchmark data sets and data sets with
problem-specific knowledge for the instances' importance.

\begin{table}[!th]
\centering 
  \begin{tabular}{@{\extracolsep{\fill}} l S[table-format=3.2] S[table-format=3.2]  S[table-format=3.2] }
  \hline\noalign{\smallskip}
    \multicolumn{1}{l}{} & 
    \multicolumn{1}{c}{Training set} &
    \multicolumn{1}{c}{Testing set}  &
    \multicolumn{1}{c}{Feature} \\
    \multicolumn{1}{l}{Data set} & 
    \multicolumn{1}{c}{size} &
    \multicolumn{1}{c}{size}  &
    \multicolumn{1}{c}{size} \\ 
  \noalign{\smallskip}\hline\noalign{\smallskip}
  breast cancer & 100 & 783  & 10  \\
  \noalign{\smallskip}
  mushrooms & 200 & 7924  & 112  \\
  \noalign{\smallskip}
  waveform & 400 & 4600  & 21 \\
  \noalign{\smallskip}
  banana & 400 & 4900  & 2  \\
  \noalign{\smallskip}
  skin nonskin & 50 0& 244557 & 3  \\
  \noalign{\smallskip}
  splice & 1000 & 2175 & 60 \\
  \noalign{\smallskip}
  image & 1300 & 1010  & 18  \\
  \noalign{\smallskip}
  adult &  1605 & 30956 & 123  \\
  \noalign{\smallskip}
  MNIST 2's vs. 9's &  11907  & 2041  & 780 \\
  \noalign{\smallskip}
  MNIST 1's vs. 7's &  13007 & 2163  & 780 \\
  \noalign{\smallskip}
  MNIST 3's vs. 6's & 12049  & 1968  & 780 \\
  \noalign{\smallskip}
  MNIST 0's vs. 8's &  11774  & 1954  & 780 \\ 
  \noalign{\smallskip}
  Parkinson's speech & 520 & 520  & 26 \\
  \noalign{\smallskip}
  bank marketing & 300 & 40888  & 11 \\
  \noalign{\smallskip}\hline
\end{tabular}
\caption{Overview of the data sets.}
\label{table:datasets}
\end{table}

\subsubsection{Benchmark data sets}
\label{subsubsec:benchmark_datasets}
These are commonly-used benchmark data sets from various repositories:
breast cancer, mushrooms, waveform, banana, skin nonskin, splice,
image and adult \citep{Lichman2013, mldata, delve}.  Also, we use the
MNIST data set \citep{Lecun1998} to generate four binary
classification problems: 2's vs. 9's; 1's vs. 7's; 3's vs. 6's; and
0's vs. 8's.

\subsubsection{Data sets with importance scores for training instances}
\label{subsubsec:importance_datasets}
Apart from the standard benchmark data sets, we use two data sets that
provide additional information for assigning the individual importance
of each training instance. These data sets are:
\begin{itemize}

\item \emph{Parkinson's speech data set}. The data set is provided by
  \cite{Sakar2013}. The instances consist of 26 features extracted
  using multiple types of sound recordings collected by 20 healthy and
  20 PWP (People with Parkinson's disease) individuals. Using this
  data set we build binary models to predict if a new measurement is
  from a healthy or PWP individual.

  Each of the 40 individuals in the provided training set has 26
  recordings; a total of 1040 recordings is given. Because the
  provided testing set consists of 168 recordings only from PWP
  individuals, we modified the training/testing set splits for our
  experiments. The individuals are split into training and testing
  sets so that each set has measurements originating from 10 healthy
  and 10 PWP individuals. The training set contains the individuals
  with IDs: 1--10, 21--30; the testing set contains the individuals
  with IDs: 11--20 and 31--40.

  Also, the data set provides the UPDRS (Unified Parkinson's disease
  rating scale) score for each PWP individual.  The UPDRS score is an
  indication for the severity of the disease and it is determined by
  an expert via interview and clinical observation
  \citep{Goetz2008}. In Experiment \#2, we will use the UPDRS score to
  generate instance weights $C_i$.

\item \emph{Bank marketing data set}. The data set is provided by
  \cite{Moro2014}. It contains data of phone call marketing campaigns
  from a Portuguese banking institution. The data set consists of
  $41188$ instances. In our experiments, we use the following
  numerical features: age, previous numbers of calls, employment
  variation rate, consumer price index, consumer confidence index,
  euribor 3 month rate, number of employees; and the following boolean
  features: ``has credit in default?'', ``has housing loan?'', ``has
  personal loan?'', ``is married?''. The class of the instances is
  whether a client subscribed to the product of the campaign or not.

  The data set also provides the duration of the phone calls. The goal
  is to generate the models that predict the outcome of the phone call
  before it is actually made, thus, since the phone call duration is
  not known beforehand, it should not be used as a feature
  \citep{Moro2014}. Nevertheless, we will take advantage of this
  information in Experiment \#2 to generate instance weights, $C_i$,
  for weighted SVM training.

\end{itemize}

\subsection{Experiment \#1: Selecting optimal class weights $C^{+}/C^{-}$}
\label{subsec:cplusminus}
A very common application of weighted SVM is the use of class weights
that alter the contribution of each class to the training error. That
is, the weights are assigned as: $C_i = C^+$ if $y_i = +1$ and $C_i =
C^-$ if $y_i = -1$. In fact, this weighting scheme is so common that
most SVM solver libraries provide it as part of the ``standard'' SVM
functionality.

Following the hyperparameter selection paradigm described in Section
\ref{sec:weight_optimization}, we execute a grid search for the
optimal training parameters for the values of $C^+$ and $C^-$. The
parameters $C^+$ and $C^-$ are taking values from $log_{2}C^{\pm} =
-6$ to $log_{2}C^{\pm} = +10$ with logarithmic step $0.5$. This grid
search results to $J = 1089$ weighted SVM models for each data set.

\begin{table}[!t]
  \renewrobustcmd{\bfseries}{\fontseries{b}\selectfont}
  \renewrobustcmd{\boldmath}{}
 \centering
\begin{tabular*}{\textwidth}{@{\extracolsep{\fill}} llllllll}
    \hline\noalign{\smallskip}
    \multicolumn{1}{l}{Data set} & &
    \multicolumn{1}{l}{Span-rule} & 
    \multicolumn{1}{l}{Span bound} & 
    \multicolumn{1}{l}{$K$-fold CV} &
    \multicolumn{1}{l}{$\xialpha$ bound} & &
    \multicolumn{1}{l}{Min. error} \\
    \noalign{\smallskip} \cline{1-1} \cline{3-6} \cline{8-8} \noalign{\smallskip}
    breast cancer & & \bf{0.0360} & 0.0395  & 0.0377 & \bf{0.0360} & & 0.0292 \\
    \noalign{\smallskip}
    mushrooms & & \bf{0.0151} &0.0154 & 0.0456 & 0.0623 & & 0.0133 \\
    \noalign{\smallskip}
    waveform & & 0.1141 & 0.1226 & \bf{0.1104} & 0.1422 & & 0.1052 \\
    \noalign{\smallskip}
    banana & & \bf{0.2324} & 0.4476 & \bf{0.2324} & 0.4476 & & 0.2324\\
    \noalign{\smallskip}
    skin nonskin & & \bf{0.0086} & 0.0143 & 0.0131 & 0.0104 & &  0.0061 \\
    \noalign{\smallskip}
    splice & & \bf{0.1080} & 0.1237 & 0.1085 & 0.1591 & & 0.1016 \\
    \noalign{\smallskip}
    image & & \bf{0.0623} & 0.1158 & \bf{0.0623} & 0.0852 & & 0.0623 \\
    \noalign{\smallskip}
    adult & & \bf{0.1585} & 0.2405 & 0.1628 & 0.2405 & & 0.1557 \\
    \noalign{\smallskip}
    MNIST 2's vs. 9's & & \bf{0.0049} & 0.0103 & 0.0059 & 0.0103 & & 0.0049 \\
    \noalign{\smallskip}
    MNIST 1's vs. 7's & & \bf{0.0042} & 0.0065 & \bf{0.0042} & 0.0065  & & 0.0032\\
    \noalign{\smallskip}
    MNIST 3's vs. 6's & & \bf{0.0020} & 0.0030 & 0.0030 & 0.0091 & & 0.0020 \\
    \noalign{\smallskip}    
    MNIST 0's vs. 8's & & 0.0067  & \bf{0.0061} & \bf{0.0061} & 0.0092 & & 0.0051 \\
    \noalign{\smallskip}\hline
  \end{tabular*}
  \caption{Results for the hyperparameter selection experiment for
    class weights $C^+/C^-$ (\emph{Experiment \#1}). The error on the
    testing set for the model selected by each hyperparameter
    selection method is reported. Best performing method for each data
    set is marked with bold font. For reference, last column presents
    the performance of the model with the minimum error on the test
    set.}
  \label{table:Cpm_results}
\end{table}

The RBF kernel, $K(\x, \x') = e^{-\gamma||\x-\x'||^2}$, is used. RBF's
parameter $\gamma$ is usually selected from a hyperparameter selection
procedure or may be assigned through heuristics. In order to limit the
computational effort, and since our primary goal is to compare the
performance of the hyperparameter selection methods with diverse
instance weights, we use a commonly followed heuristic: each feature
dimension is linearly scaled in $[0, 1]$ and the value of $\gamma$ is
set to $\frac{1}{d}$, where $d$ is the dimensionality of the feature
space (that is, $\x_i \in \mathbb{R}^d$).

The $12$ standard benchmark data sets of Section
\ref{subsubsec:benchmark_datasets} are used for this evaluation.
Table \ref{table:Cpm_results} presents the error on the testing sets
for the weighted SVM model selected by each hyperparameter selection method
as the optimal one for each data set. If the minimum of a prediction
rule is encountered more than once, the worst case outcome is
reported. For reference, we also provide the test error of the model
that exhibited the best performance.

The results demonstrate that in $10$ out of $12$ data sets the
span-rule performed better or equal to $K$-fold CV and selected models
with performance close to the minimum possible error. The span bound
and the the $\xialpha$ bound demonstrated good performance in some
data sets but, in general, their performance was inferior to span-rule
and $K$-fold CV.

\subsection{Experiment \#2: Selecting individual instance weights, $C_i$, from scores, $q_i$ }
\label{subsec:weight_selection}
In this experiment we generate sets of individual weights $C_i$ using
the problem-specific knowledge for the importance of each instance. We
quantify the importance of each instance by assigning a score
(membership) value $q_i$. Larger $q_i$ values are assigned to the
instances that should be considered more important during
training. Thus, the goal of the hyperparameter selection methods is to
identify the optimal mapping of the scores, $\{q_i\}$, to sets of
instance weights, $\{C_i\}$.

The Parkinson's speech and the bank marketing data set are used for
this set of experiments (see Section
\ref{subsubsec:importance_datasets}). For the Parkinson's speech data
set, the UPDRS score describes the severity of the disease symptoms
and, thus, the importance of a measurement for the PWP class. To this
end, the UPDRS scores are scaled in $[0,1]$ and used as the scores,
$q_i$, for the training instances originating from PWP
individuals. The scores for the training instances from the healthy
individuals are set to $1$.

For the bank marketing data set, the score, $q_i$, of each instance is
set to the duration of the phone call scaled in $[0,1]$. The call
duration should not be used as direct input for the classifiers in
order to be able to use the resulting model in real applications
\citep{Moro2014}; however, it can be used to incorporate the available
problem-specific information in the form of instance weights during
model training. For example, calls that last zero or a few seconds are
considered less informative and their effect can be suppressed using
smaller instance weights. On the other hand, the calls that last
longer disclose more information regarding successful or unsuccessful
product subscriptions.

For these two data sets, the mapping from $q_i$ to $C_i$ is performed
by a family of sigmoidal functions parameterized by $A,B,C$ in the
form:
\begin{equation}
\label{eq:sigmoidal}
C_i(q_i; A,B,C) = \max\left(\frac{C}{1+e^{-A \cdot (q_i - B)}}, \sigma\right), i = 1, \dots, l
\end{equation}
where $\sigma$ is a small value introduced for training stability (in
our experiments $\sigma = 0.01$).

A member of the mapping function family is defined from a combination
for the values of $A$, $B$ and $C$.  Again, following the
hyperparameter selection paradigm described in Section
\ref{sec:weight_optimization}, we execute a grid search for the
optimal $(A,B,C)$ combination: the parameter $A$ is taking values from
$1$ to $10$ with step $1$; the parameter $B$ is taking values from $0$
to $0.9$ with step $0.1$; and the parameter $C$ is taking values from
$log_{2}C = -6$ to $log_{2}C = +10$ with logarithmic step $1$. This
grid search results to $J = 1700$ weighted SVM models.

Table \ref{table:Ci_results} presents the error on the testing set
for the model selected by each hyperparameter selection method as the
optimal one for each data set. For reference, we also provide the
testing error of the model that exhibited the best performance on the
testing set. Span-rule performed better in both data sets
outperforming the other hyperparameter selection methods.

\begin{table}[!t]
  \renewrobustcmd{\bfseries}{\fontseries{b}\selectfont}
  \renewrobustcmd{\boldmath}{}
 \centering
\begin{tabular*}{\textwidth}{@{\extracolsep{\fill}} llllllll}
  \hline\noalign{\smallskip}
    Data set & &
    Span-rule & 
    Span bound & 
    $K$-fold CV &
    $\xialpha$ bound & &
    Min. error \\
    \noalign{\smallskip} \cline{1-1} \cline{3-6} \cline{8-8} \noalign{\smallskip}
    Parkinson's speech & & \bf{0.3865} & 0.5000 & 0.4000 & 0.5134 & & 0.3692 \\
    \noalign{\smallskip}
    bank marketing & & \bf{0.2218} & 0.2801 & 0.2297 & 0.2803 & & 0.2218 \\
    \noalign{\smallskip}\hline
  \end{tabular*}
  \caption{Results for the hyperparameter selection experiment with
    individual weights $C_i$ (\emph{Experiment \#2}). The error on the
    testing set for the model selected by each hyperparameter
    selection method is reported. Best performing method for each data
    set is marked with bold font. For reference, last column presents
    the performance of the model with the minimum error on the test
    set.}
\label{table:Ci_results}
\end{table}

Overall, the results of Experiments \#1 and \#2 demonstrate that the
span-rule is a very effective and reliable hyperparameter selection
method for the weighted SVM algorithm. In almost all $14$ data sets in
both experimental settings, the span-rule demonstrated better or equal
performance compared to all other methods.

\subsection{Experiment \#3: Predicting the test error value using span-rule}
\label{subsec:predicting_curves}
The primary quality we are looking for in a hyperparameter selection
method is to be able to discover the model with the best possible
generalization performance. Experiments \#1 and \#2 demonstrated that
span-rule is the best method for this task. In addition, a good
hyperparameter selection method should also predict the value of the
actual test error.

In this experiment, we examine the performance of the hyperparameter
selection methods on predicting the actual value of the testing
error. We evaluate the span-rule, the $K$-fold CV and the $\xialpha$
bound but we excluded the span bound of Theorem 1 since the latter is
not bounded above (that is, it can be greater than 1).

\begin{table}[!b]
  \renewrobustcmd{\bfseries}{\fontseries{b}\selectfont}
  \renewrobustcmd{\boldmath}{}
 \centering
\begin{tabular}{lllll}
  \hline\noalign{\smallskip}
  \multicolumn{1}{l}{Data set} & 
  \multicolumn{1}{l}{Span-rule} &
  \multicolumn{1}{l}{$K$-fold CV} &
  \multicolumn{1}{l}{$\xialpha$ bound} \\
  \noalign{\smallskip}\hline\noalign{\smallskip}
  breast cancer & \bf{0.0229} & 0.0494 & 0.0419 \\
  \noalign{\smallskip}
  mushrooms & \bf{0.0151} & 0.0384 & 0.1046 \\ 
  \noalign{\smallskip}
  waveform & \bf{0.0256} & 0.0271 & 0.0912  \\
  \noalign{\smallskip}
  banana & 0.0150 & \bf{0.0128} & 0.2150  \\
  \noalign{\smallskip}
  skin nonskin & \bf{0.0204} & 0.0496 & 0.0884 \\
  \noalign{\smallskip}
  splice & \bf{0.0156} & 0.0199 & 0.2135 \\
  \noalign{\smallskip}
  image & \bf{0.0134} & 0.0213 & 0.1251 \\
  \noalign{\smallskip}
  adult & \bf{0.0066} &  0.0136 & 0.1495 \\
  \noalign{\smallskip}
  MNIST 2's vs. 9's & \bf{0.0108} & 0.0229 & 0.0539 \\
  \noalign{\smallskip}
  MNIST 1's vs. 7's & \bf{0.0060} & 0.0198 & 0.0404 \\
  \noalign{\smallskip}
  MNIST 3's vs. 6's & \bf{0.0068} & 0.0188 & 0.0389 \\
  \noalign{\smallskip}
  MNIST 0's vs. 8's & \bf{0.0077} & 0.0182 & 0.0420 \\
  \noalign{\smallskip}
  Parkinson's speech & 0.0152 & \bf{0.0134} & 0.2167 \\
  \noalign{\smallskip}
  bank marketing & \bf{0.0163} & 0.0824 &  0.1730 \\
  \noalign{\smallskip}\hline
\end{tabular}
\caption{Results for {\it Experiment \#3}. Table reports the RMSE
  between the methods' predictions and the actual test error. The
  results are calculated using the weighted SVM models trained for
  each data set in {\it Experiments \#1} (1089 models per data set)
  and {\it Experiment \#2 } (1700 models per data set). Best
  performing method for each data set is marked with bold font.}
  \label{table:correlation}
\end{table}

Table \ref{table:correlation} shows the root-mean-square error (RMSE)
between the error predictions and the actual test error for the $1089$
weighted SVM models trained for each data set in Experiment \#1 and
the $1700$ weighted SVM models trained for each data set in Experiment
\#2.

Span-rule demonstrates the best RMSE in $12$ of the $14$ data
sets. Span-rule and $K$-fold CV perform fairly similarly and both
exhibit low RMSE. On the other hand, the $\xialpha$ bound does not
demonstrate consistently the desired trait. The ability to predict the
value of the test error is better illustrated using figures, as shown
next.

\begin{figure}[!h]
\centering
\subfloat[Over $C^+$]{\includegraphics[width=0.45\textwidth]{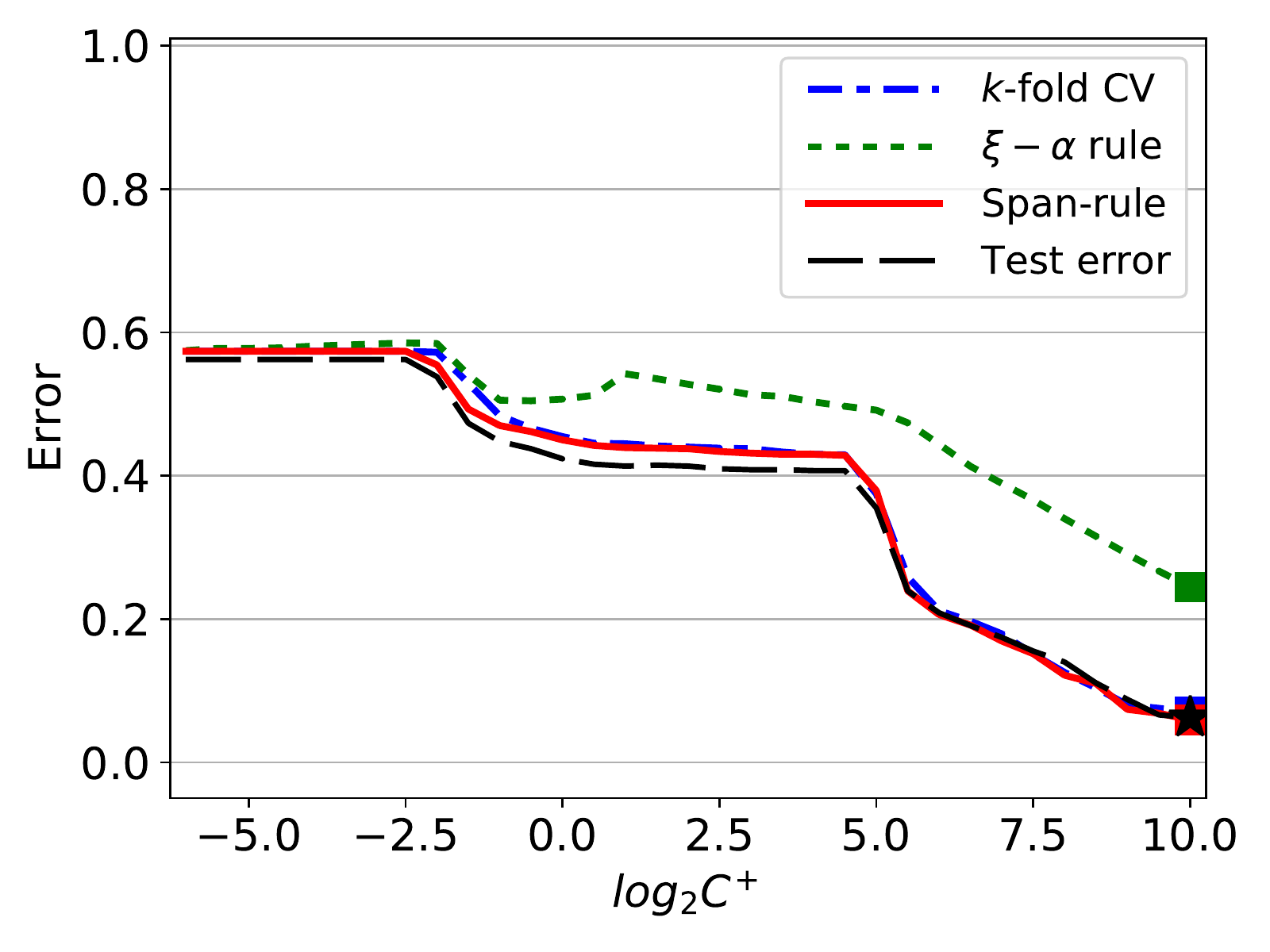}}%
\subfloat[Over $C^-$]{\includegraphics[width=0.45\textwidth]{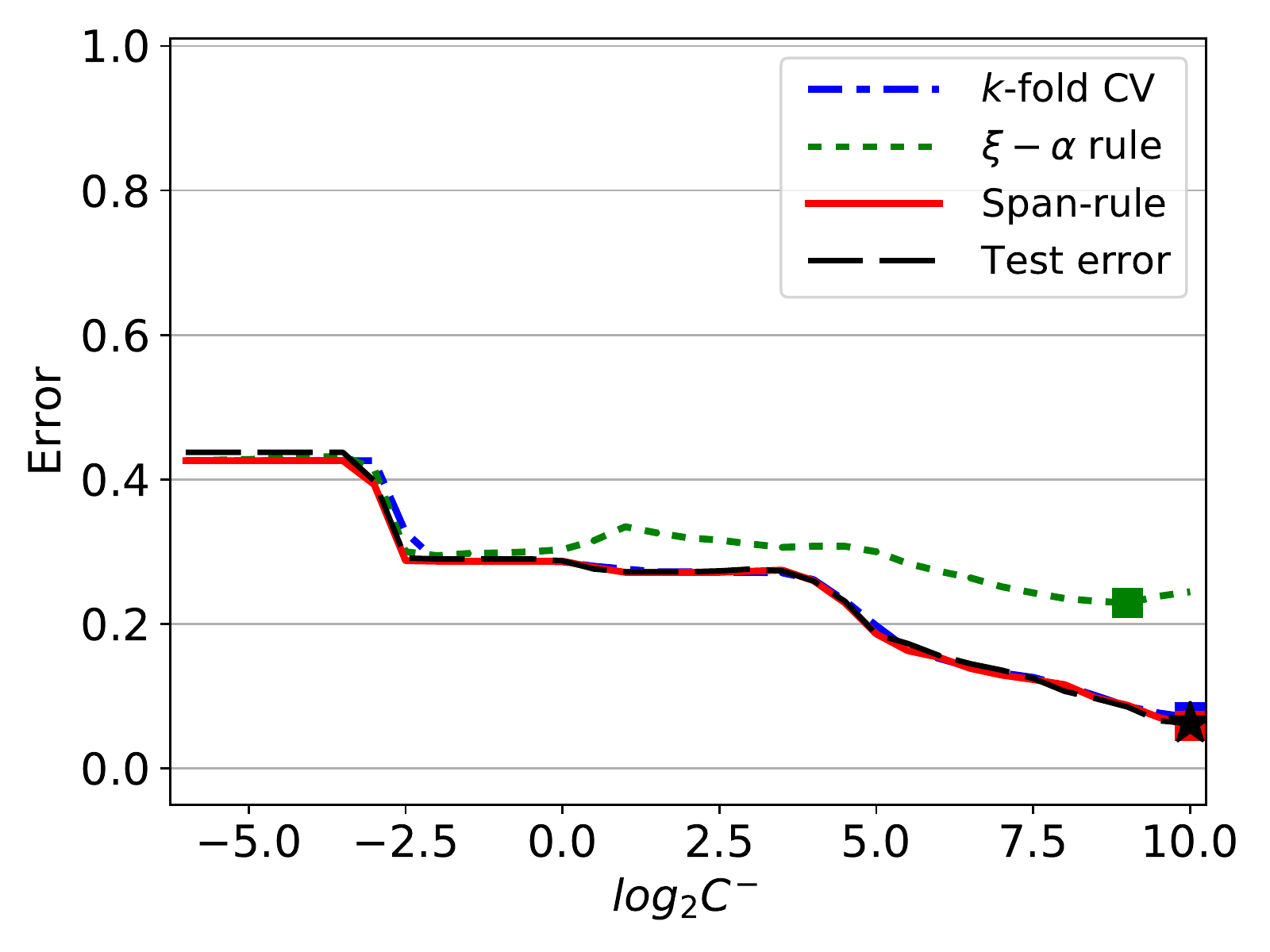}}%
\caption{Image data set. Test error prediction curves and test error
  curve over each training parameter ({\it Experiment \#3}). The
  minimum for each prediction method is marked with a colored
  ``$\square$''; the minimum found test error is marked with a
  ``$\star$''.}
\label{fig:image}
\end{figure}
\begin{figure}[!h]
\centering
\subfloat[Over $C^+$]{\includegraphics[width=0.45\textwidth]{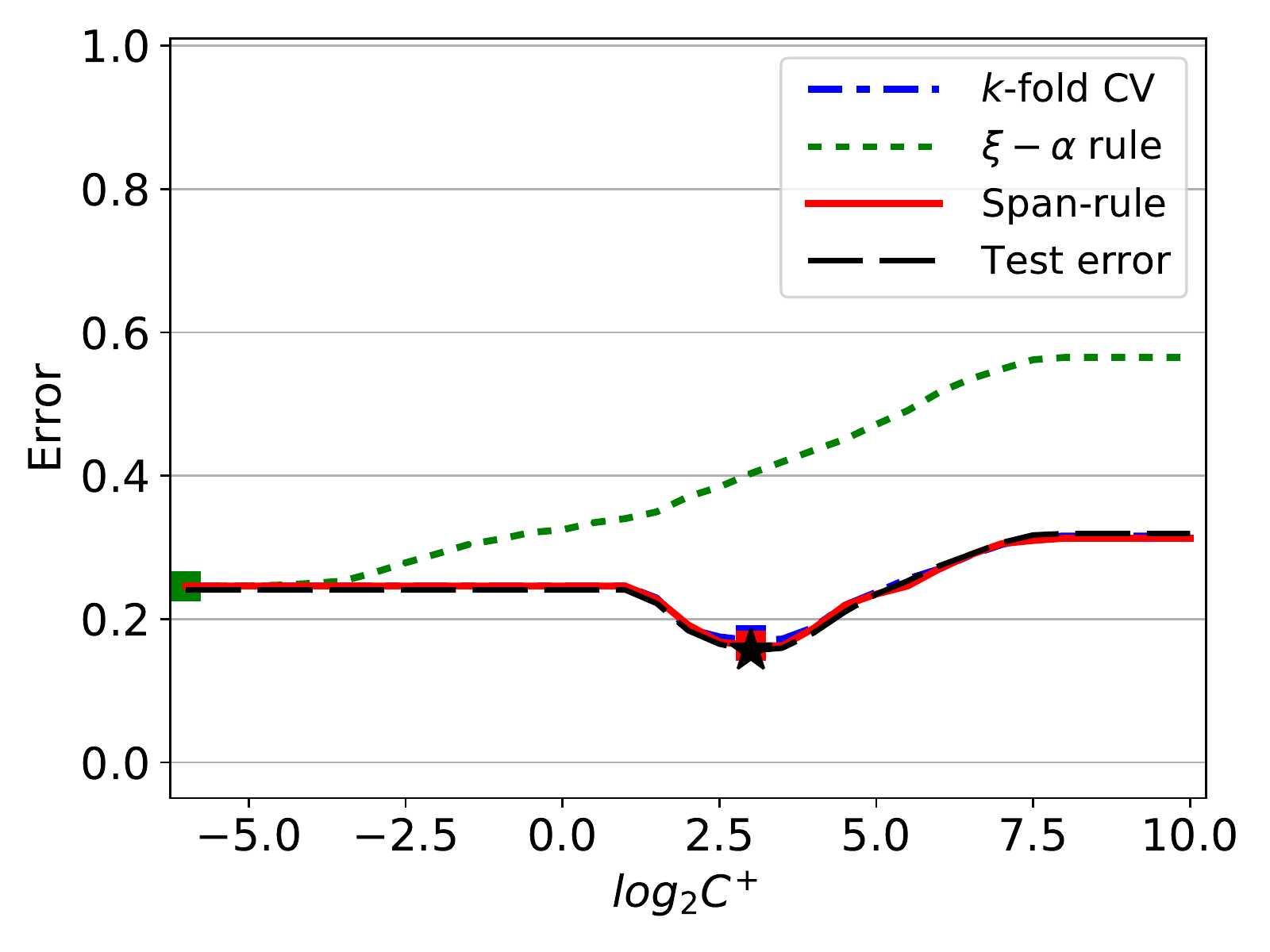}}%
\subfloat[Over $C^-$]{\includegraphics[width=0.45\textwidth]{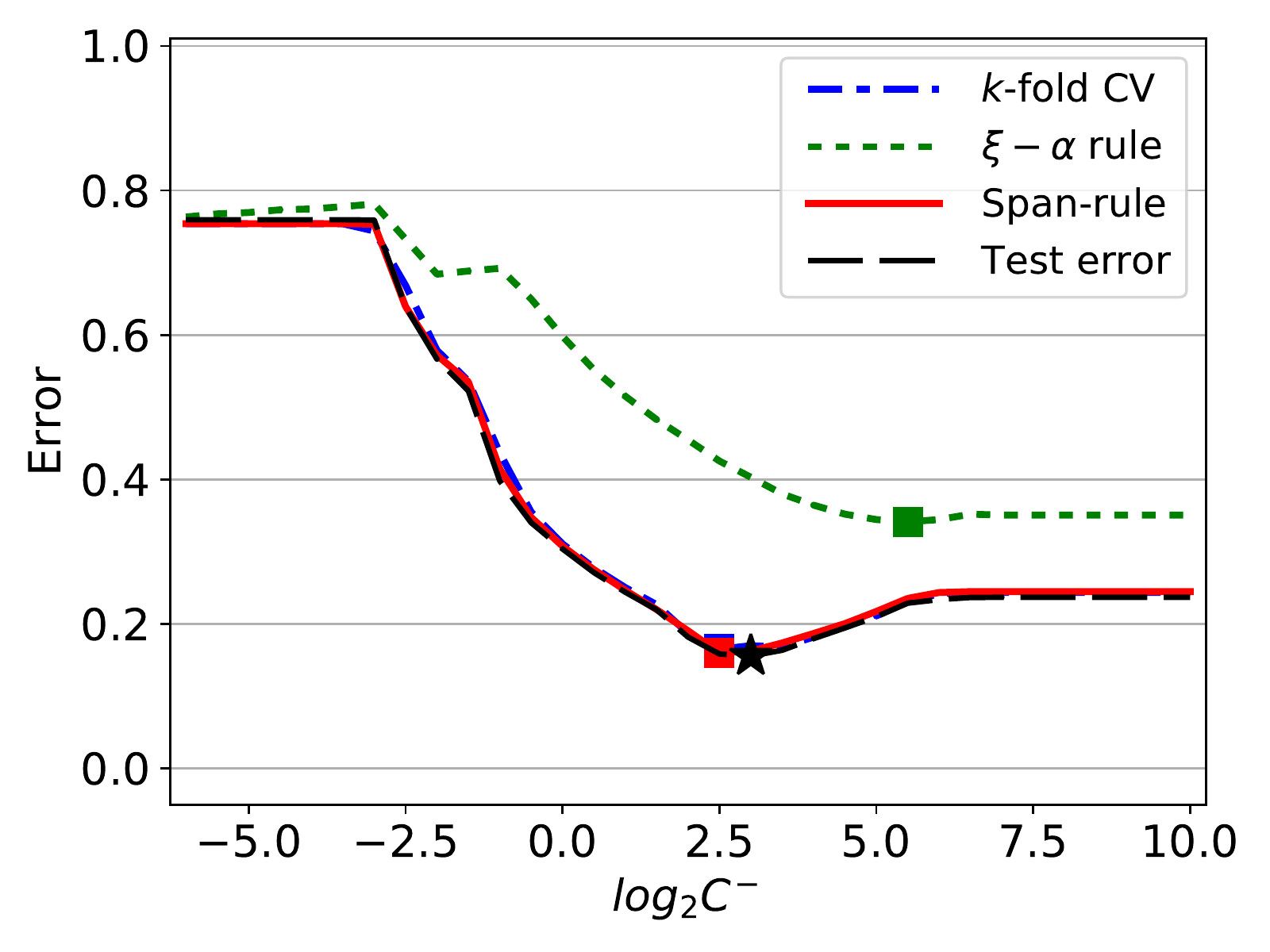}}%
\caption{Adult data set. Test error prediction curves and test error
  curve over each training parameter ({\it Experiment \#3}). The
  minimum for each prediction method is marked with a colored
  ``$\square$''; the minimum found test error is marked with a
  ``$\star$''.}
\label{fig:adult}
\end{figure}
\begin{figure}[!h]
\centering
\subfloat[Over $A$]{\includegraphics[width=0.36\textwidth]{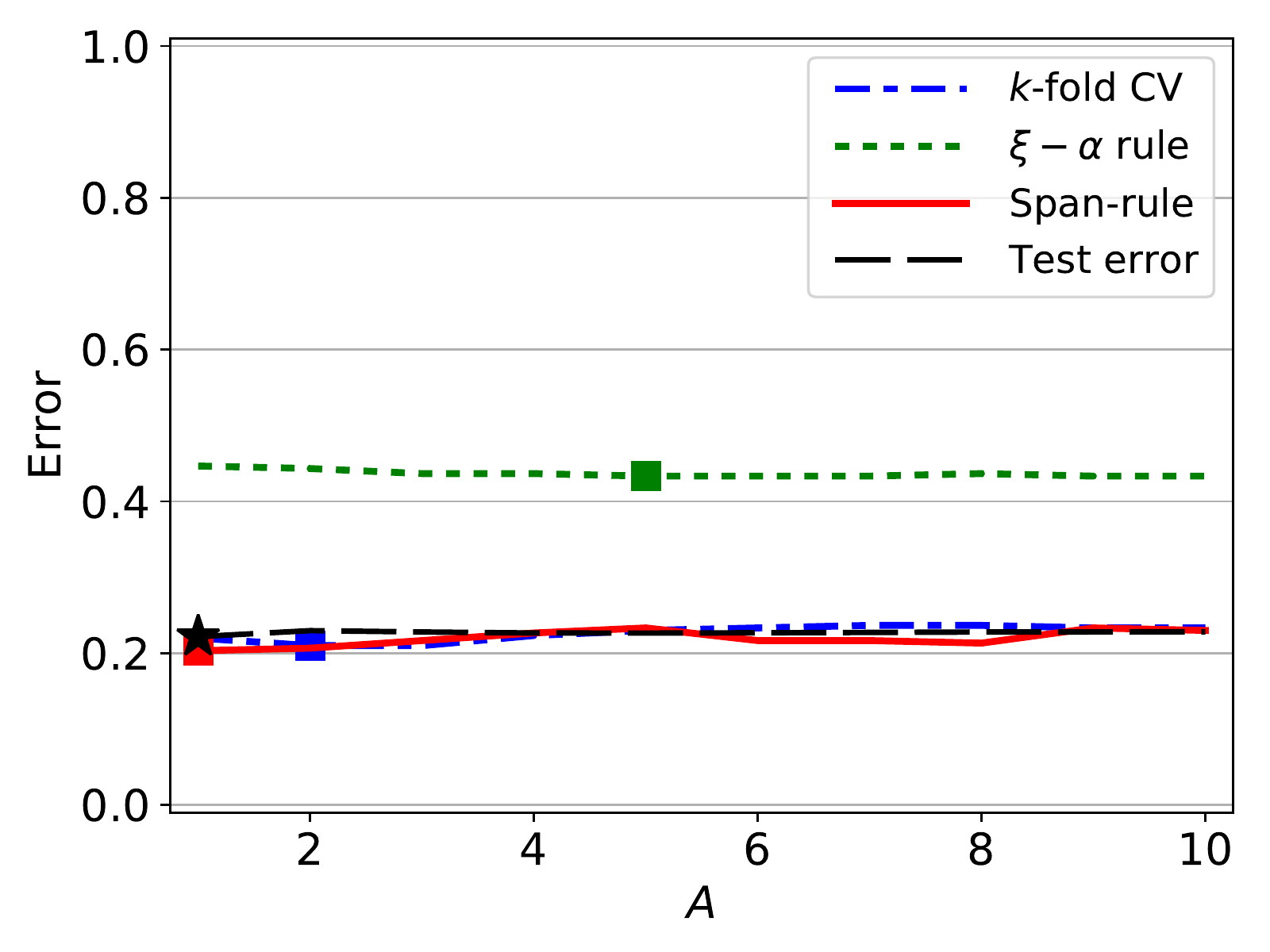}}%
\subfloat[Over $B$]{\includegraphics[width=0.36\textwidth]{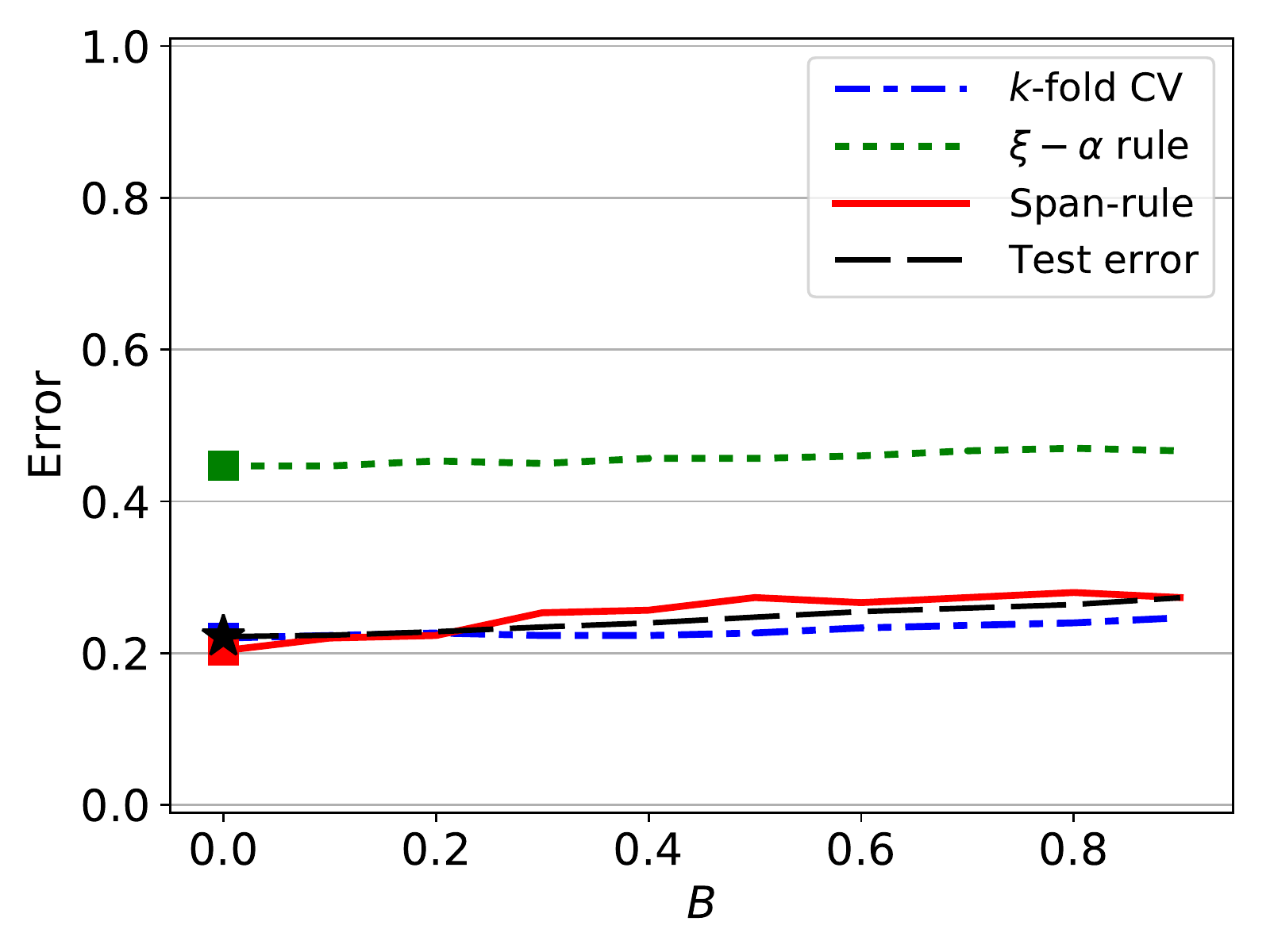}}\\
\subfloat[Over $C$]{\includegraphics[width=0.36\textwidth]{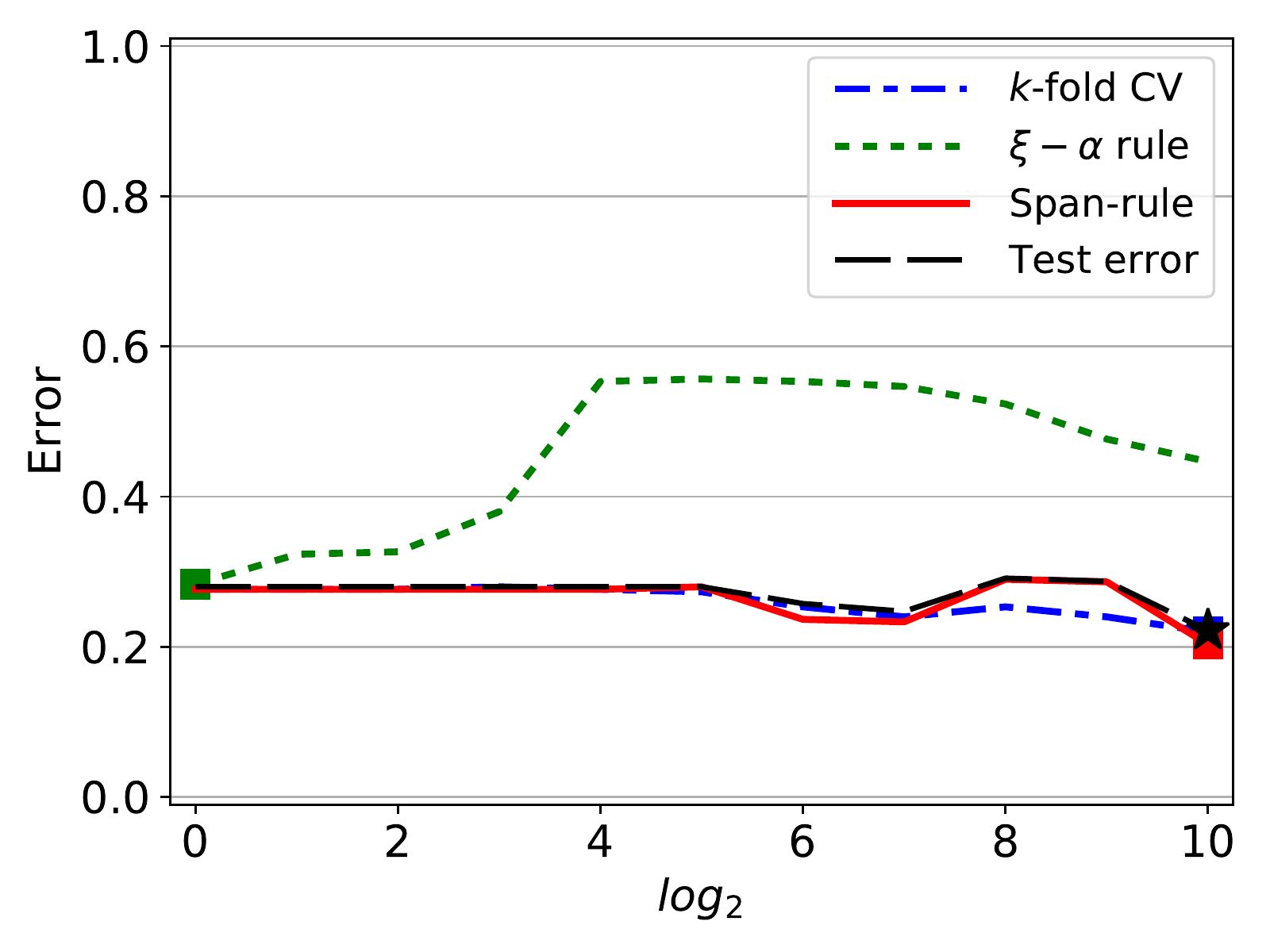}}%
\caption{Bank marketing data set. Test error prediction curves and
  test error curve over each training parameter ({\it Experiment
    \#3}). The minimum for each prediction method is marked with a
  colored ``$\square$''; the minimum found test error is marked with a
  ``$\star$''.}
\label{fig:bank}
\end{figure}

Figures \ref{fig:image}, \ref{fig:adult}, and \ref{fig:bank} show the
test error prediction curves and the test error curve for the image,
adult, and bank marketing data sets respectively. In our experiments
the weighted SVM models depend on the grid combinations of two
training parameters, $(C^+, C^-)$, in Experiment \#1 and three
training parameters, $(A, B, C)$, in Experiment \#2. In each sub-plot
the curves are plotted against one of the training parameters whereas
the values of the other(s) are set to the one(s) that gave the model
with the minimum actual test error.

The graphs illustrate that the span-rule for weighted SVM is able to
accurately predict the value of the test error over varying
parameterizations. Most importantly, span-rule selects parameters that
belong in the area around the parameters that yield the minimum test
error. Regarding the other hyperparameter selection methods, the
$K$-fold cross-validation also performs well. On the other hand, the
$\xialpha$ bound is very conservative and fails to predict the value,
or even the trend, of the actual test error curve.

\subsection{Experiment \#4: Is the value of span, $S_p$, defined most of the time?}
\label{subsec:existence_Lp}
Previous experiments demonstrated that the span-rule for weighted SVM
is an effective method for hyperparameter selection and test error
prediction. Span-rule is a corollary of Theorem 2 that guarantees $\Lp
\neq \emptyset$ and, thus, the value of $S_p$ can always be calculated
under theorem's assumptions.

Nevertheless, it is important to investigate how often the set $\Lp$
is indeed non-empty by examining how often the inequality of Lemma 1
is satisfied. Table \ref{table:feasibilities} shows the average number
of $\Lp = \emptyset$ occurrences and, for reference, the average
number of support vectors per weighted SVM model. The displayed values
are averaged over the $1089$ weighted SVM models trained for each data
set in Experiment \#1 and the $1700$ weighted SVM models trained for
each data set in Experiment \#2. 

Overall, the experiments demonstrate that the occurrence of empty
$\Lp$ is a very rare one. Notably, there is less than one $\Lp =
\emptyset$ occurrence on average for each weighted SVM model and this
behavior is consistent across data sets.

\begin{table}[!t]
 \centering
\begin{tabular}{@{\extracolsep{\fill}} l 
    S[table-format=1.3,table-figures-uncertainty=1] S[table-format=1.1,table-figures-uncertainty=1] S[table-format=1.1,table-figures-uncertainty=1]}
  \hline\noalign{\smallskip}
  \multicolumn{1}{l}{Data set} & 
    \multicolumn{1}{c}{$\Lp = \emptyset$ occurrences} &
    \multicolumn{1}{c}{In-bound SVs ($n^*$)} & 
    \multicolumn{1}{c}{Total SVs ($n$)} \\
    \noalign{\smallskip}\hline\noalign{\smallskip}
    breast cancer & 0.128 \pm 0.461 & 4.6 \pm 2.0 & 33.3 \pm 22.6  \\
    \noalign{\smallskip}
    mushrooms &  0.0 \pm 0.0 & 27.6 \pm 10.1 & 93.0 \pm 35.9 \\ 
    \noalign{\smallskip}
    waveform & 0.003 \pm 0.068 & 16.0 \pm 7.2 & 188.7 \pm 58.3 \\
    \noalign{\smallskip}
    banana & 0.0 \pm 0.0 & 17.2 \pm 14. & 277.3 \pm 47.1 \\
    \noalign{\smallskip}
    skin nonskin & 0.197 \pm 0.621 & 7.3 \pm 3.8 & 167.9 \pm 115.3 \\
    \noalign{\smallskip}
    splice & 0.0 \pm 0.0 & 148.2 \pm 122.9  & 581.3 \pm 78.0  \\
    \noalign{\smallskip}
    image & 0.004 \pm 0.074 &  18.8 \pm 7.0 & 640.9 \pm 160.2  \\
    \noalign{\smallskip}
    adult & 0.0 \pm 0.0 & 125.8 \pm 83. & 825.9 \pm 289.3 \\
    \noalign{\smallskip}
    MNIST 2's vs. 9's & 0.0 \pm 0.0 & 134.6 \pm 108.1 & 2368.5 \pm 1828.8 \\
    \noalign{\smallskip}
    MNIST 1's vs. 7's & 0.0 \pm 0.0 & 68.6 \pm 44.4 & 2210.4 \pm 2056.2 \\
    \noalign{\smallskip}
    MNIST 3's vs. 6's & 0.0 \pm 0.0 & 105.8 \pm 69.0 & 1935.2 \pm 1711.7 \\
    \noalign{\smallskip}
    MNIST 0's vs. 8's & 0.0 \pm 0.0 & 115.5 \pm 88.4 & 2028.6 \pm 1721.1 \\
    \noalign{\smallskip}
    Parkinson's speech & 0.041 \pm 0.428 & 16.1 \pm 10.4 & 416.4 \pm 45.8 \\
    \noalign{\smallskip}
    bank marketing & 0.222 \pm 0.615 & 12.5 \pm 9.4 & 250.9 \pm 51.2 \\
    \noalign{\smallskip}\hline
\end{tabular}
  \caption{Results for \emph{Experiment \#4}. Average number of $\Lp =
    \emptyset$ occurrences and average number of support vectors per
    data set. The values are calculated using the weighted SVM models
    trained for each data set in {\it Experiments \#1} (1089 models
    per data set) and {\it Experiment \#2} (1700 models per data
    set).}
  \label{table:feasibilities}
\end{table}

This finding essentially confirms the assumption of Theorem 2 which
asserts that $\Lp \neq \emptyset$ for all support vectors. Thus, in
practice, we can always safely use the span-rule estimator of
Corollary 1 by calculating the values of $S_p$ for all support vectors
of a weighted SVM model.

In addition, the results also affect our understanding for the span
bound given by Theorem 1; that is, the span bound for weighted SVM can
be as tight as the span bound for standard SVM since the value of $k$,
which represents the number of occurrences $\Lp = \emptyset$, barely
affects it.

\subsection{Experiment \#5: Efficiency evaluation of span-rule and $K$-fold CV}
\label{subsec:computational_comp}
This experiment compares the efficiency between the two most prominent
hyperparameter selection methods for weighted SVM: the span-rule and the
$K$-fold CV.

\begin{table}[!t]
  \renewrobustcmd{\bfseries}{\fontseries{b}\selectfont}
  \renewrobustcmd{\boldmath}{}
\centering
\begin{tabular}{@{\extracolsep{\fill}} l r r }
  \hline\noalign{\smallskip}
  \multicolumn{1}{l}{Data set} & 
    \multicolumn{1}{l}{Span-rule} &
    \multicolumn{1}{l}{$K$-fold CV}  \\
    \noalign{\smallskip} \hline \noalign{\smallskip}
    breast cancer & 0.031 & \bf{0.011}  \\
    \noalign{\smallskip}
    mushrooms & 0.104 & \bf{0.056} \\
    \noalign{\smallskip}
    waveform & 0.139 & \bf{0.090} \\
    \noalign{\smallskip}
    banana & 0.213 &  \bf{0.096} \\
    \noalign{\smallskip}
    skin nonskin & 0.108 & \bf{0.088} \\
    \noalign{\smallskip}
    splice & 7.637 & \bf{1.136} \\
    \noalign{\smallskip}
    image & \bf{0.416} & 0.436  \\
    \noalign{\smallskip}
    adult & 5.567 & \bf{0.990}  \\
    \noalign{\smallskip}
    MNIST 2's vs. 9's & \bf{10.423} & 123.380 \\ 
    \noalign{\smallskip}
    MNIST 1's vs. 7's & \bf{3.069} & 106.304 \\
    \noalign{\smallskip}
    MNIST 3's vs. 6's & \bf{5.116} & 108.798 \\
    \noalign{\smallskip}
    MNIST 0's vs. 8's & \bf{6.410} & 130.957  \\
    \noalign{\smallskip}
    Parkinson's speech & 0.278 & \bf{0.167} \\
    \noalign{\smallskip}
    bank marketing & 0.165 & \bf{0.060} \\
    \noalign{\smallskip}\hline
\end{tabular}
  \caption{Results for \emph{Experiment \#5}. Average execution time
    in seconds for span-rule and $K$-fold CV executions of Experiments
    \#1 (1089 executions per data set) and Experiment \#2 (1700
    executions per data set). Best average execution time is marked
    with bold font.}
  \label{table:efficiency}
\end{table}

Table \ref{table:efficiency} displays the average execution time
needed to calculate the $K$-fold CV and the values of $S_p$ for
span-rule. The models trained for the $14$ data sets of Experiments
\#1 and \#2 are used. To make the results comparable, we restricted
both LibSVM solver and the MOSEK optimizer to one processor. The
experiments were executed on a Intel Xeon E5-2650 v3 at 2.3GHz.

Span-rule demonstrates a fairly competitive execution time compared to
$K$-fold CV. In most data sets, the span-rule is slightly slower than
$K$-fold CV. Usually, such small overheads in training procedure are
not very important since, in general, the model training is performed
offline and does not affect the use of the trained model in other way.

Notably, the span-rule can be significantly faster than $K$-fold CV on
high-dimensional problems with large training sets, such as the four
MNIST data sets. To this end, the next and final experiment
investigates when is it most efficient to apply the span-rule for
weighted SVM hyperparameter selection.

\subsection{Experiment \#6: When is it faster to use span-rule than $K$-fold CV?}
\label{subsec:when_faster}
We will investigate how training set size and feature vector
dimesionality affect the efficiency of span-rule and $K$-fold CV using
synthetic data sets.

Before we delve into the details of the experiment, it is important to
understand the differences in executing the span-rule and $K$-fold CV:

\begin{itemize}

\item The $K$-fold CV requires the execution $K$ train-test procedures
  to assess the generalization performance of a given set of
  parameters. The bulk of the execution time is consumed in solving
  $K$ quadratic problems with inequality and equality constrains
  (eq. \ref{eq:fsvm_dual_obj}, \ref{eq:fsvm_dual_con1} and
  \ref{eq:fsvm_dual_con2}). Each of the $K$ quadratic problem has
  $\frac{(K-1)l}{K}$ number of unknown variables, where $l$ is the
  training set size. Due to the size of these quadratic problems,
  these are efficiently solved using the Sequential Minimal
  Optimization (SMO) algorithm \citep{platt1999fast}. Furthermore, SVM
  solvers, such as LibSVM, use a cache to store the most recently
  calculated kernel products.

\item The span-rule requires to solve $n$ quadratic problems for the
  calculation of $S_p$ with $n^*$ or $n^* + 1$ unknown variables for
  in-bound and bounded support vectors respectively; where $n$ is the
  total number of support vectors and $n^*$ is the number of in-bound
  support vectors (eq. \ref{eq:Sp_calc_1} - \ref{eq:Sp_calc_4}). In
  general, for real-word non-separable data sets, the number of
  supports vectors, $n$, is smaller than the training set size, $l$,
  and the number of in-bound support vectors, $n^*$, is a relatively
  small percentage of the total number of SVs (see also Table
  \ref{table:feasibilities}). Thus, these quadratic problems can be
  solved efficiently using general purpose Quadratic Programming
  solvers, such as the MOSEK optimizer.

\end{itemize}

\begin{figure}[!b]
\centering
\subfloat[$d = 20$]{\includegraphics[width=0.33\textwidth]{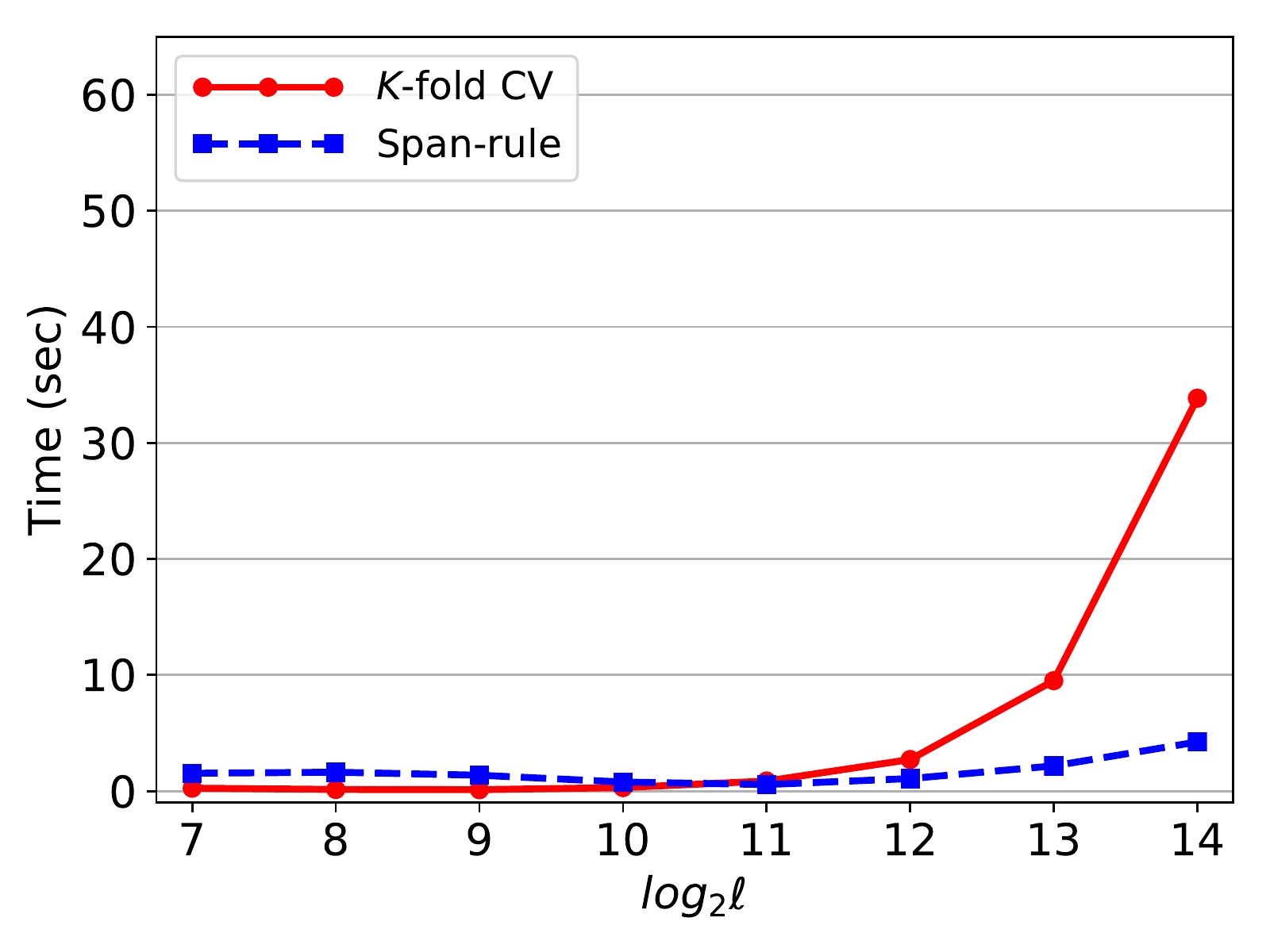}}%
\subfloat[$d = 40$]{\includegraphics[width=0.33\textwidth]{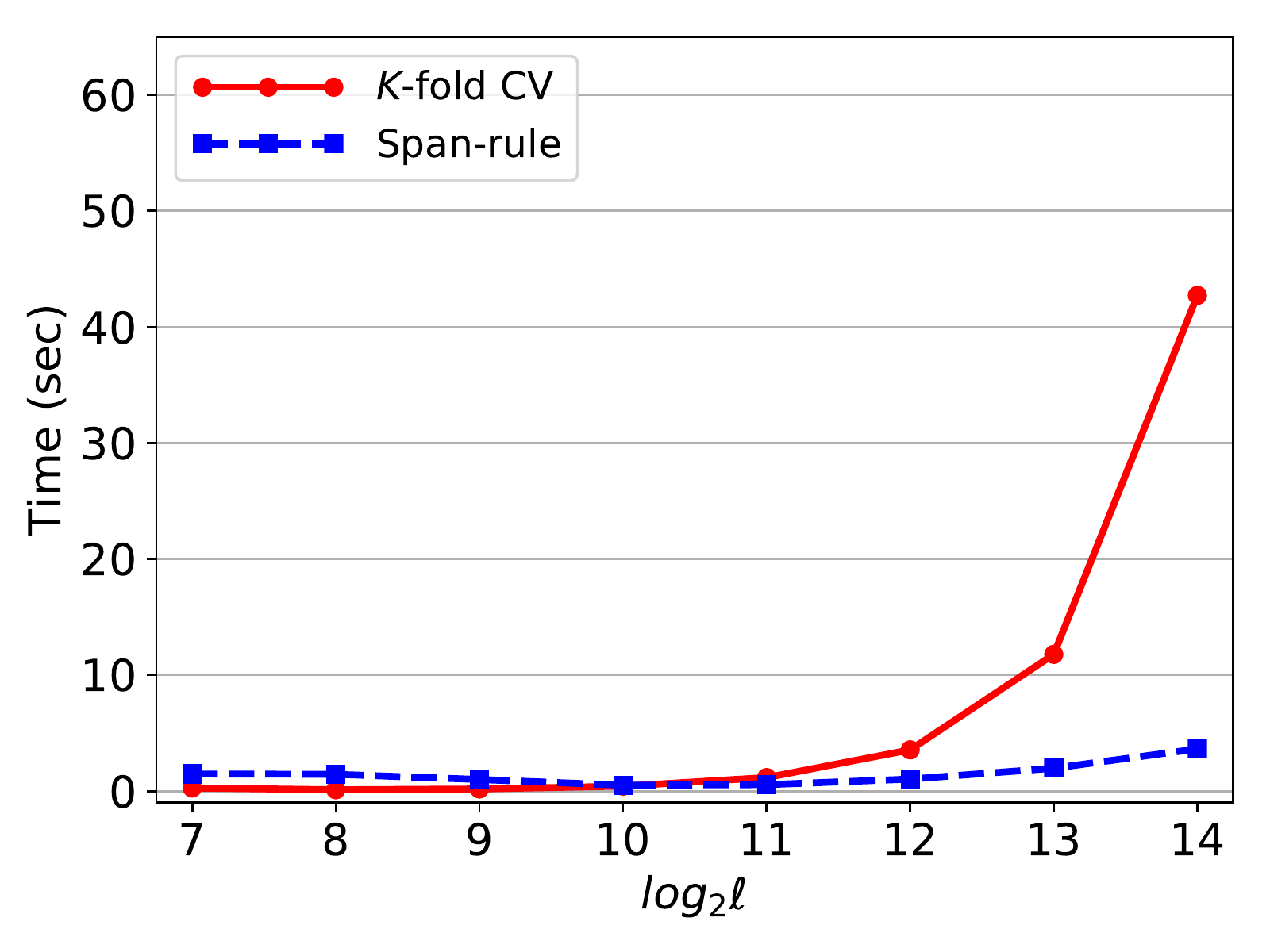}}%
\subfloat[$d = 80$]{\includegraphics[width=0.33\textwidth]{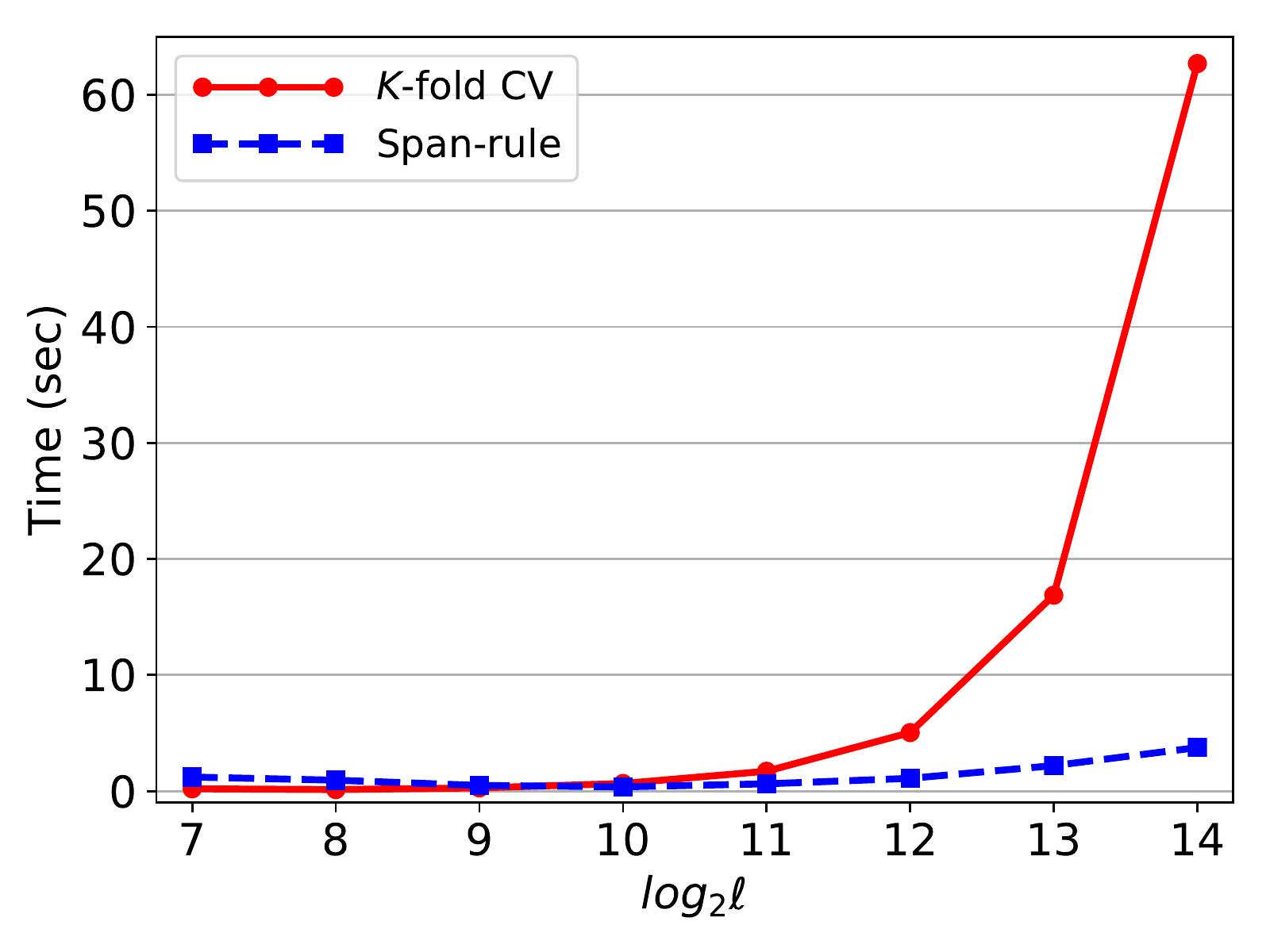}}%
\caption{Average execution time of span-rule and $K$-fold CV over
  varying training set sizes $l$ for feature dimensions $d= 20,40,80$
  ({\it Experiment \#6}).}
\label{fig:exec_time}
\end{figure}
\begin{figure}[!b]
\centering
\subfloat[$d = 20$]{\includegraphics[width=0.33\textwidth]{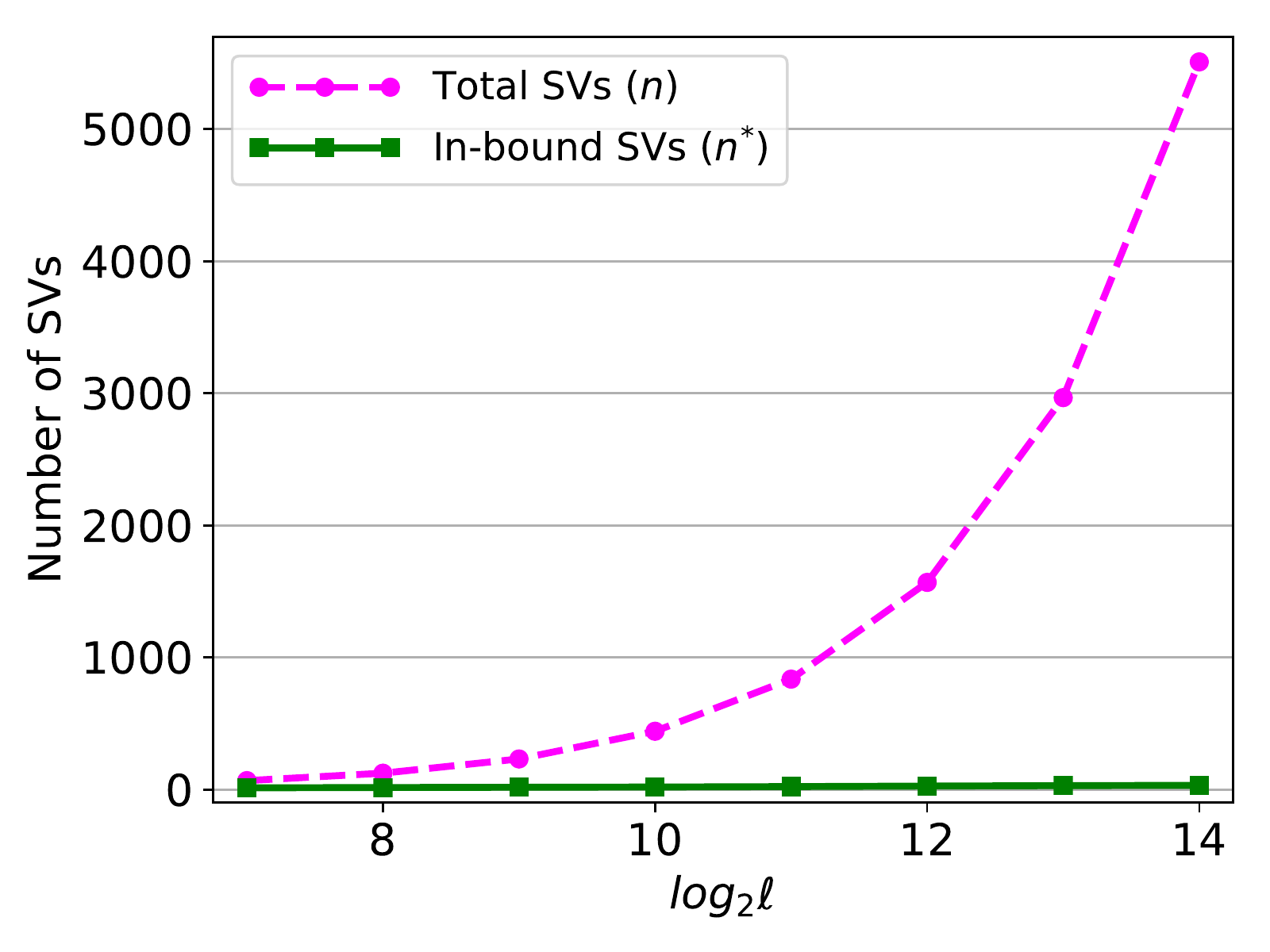}}%
\subfloat[$d = 40$]{\includegraphics[width=0.33\textwidth]{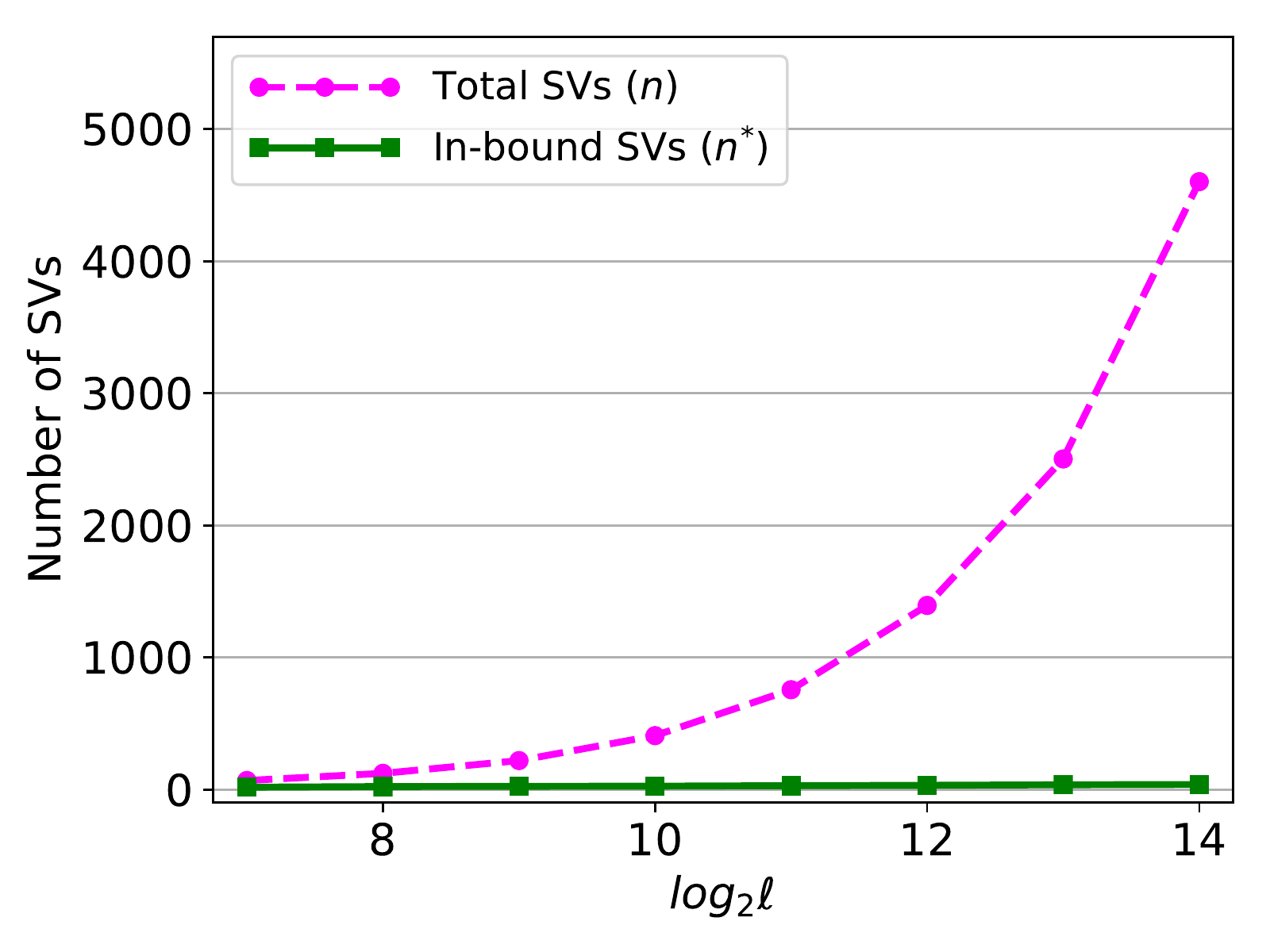}}%
\subfloat[$d = 80$]{\includegraphics[width=0.33\textwidth]{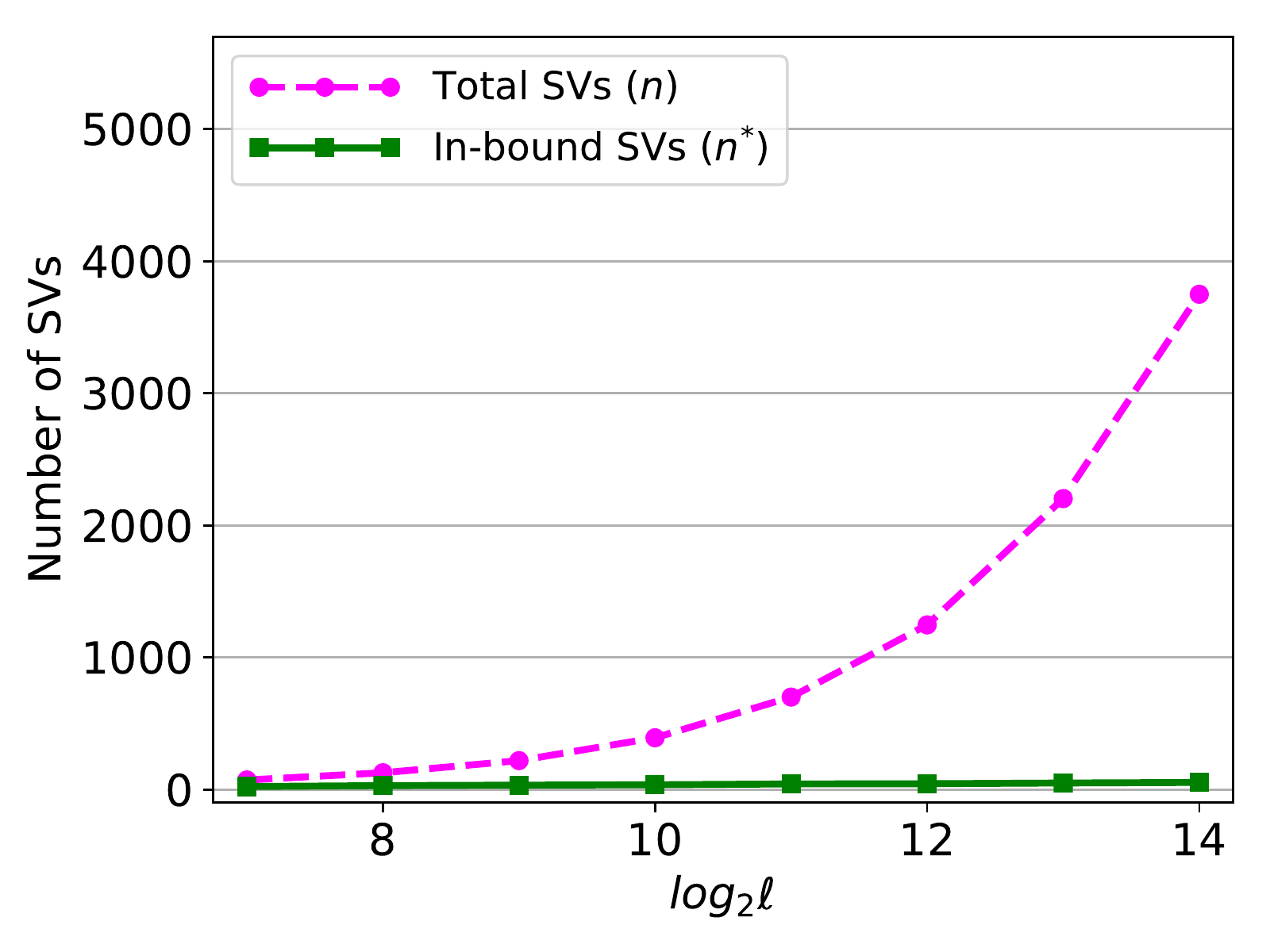}}%
\caption{Average number of in-bound and total SVs over varying
  training set sizes $l$ for feature dimensions $d= 20,40,80$ ({\it Experiment \#6}).}
\label{fig:nof_SVs}
\end{figure}

For the synthetic data sets of this experiment, we will use variations
of the ringnorm data set of \cite{breiman1996bias}. The training data
are sampled from $d$-dimensional normal distributions, $N(\x; \mu^+,
\Sigma^+)$ and $N(\x; \mu^-, \Sigma^-)$ for the positive and the
negative class respectively with priors $Pr\{y=+1\} = 0.3$ and
$Pr\{y=-1\} = 0.7$. For the positive class, we use $\mu^+ = (1, 1,
\dots, 1)$ and unit covariance matrix $\Sigma^+$; for the negative
class, we use mean $\mu^- = (0, 0, \dots, 0)$ and $\Sigma^-$ equal
four times the unit covariance matrix.

For given feature dimension $d$, we create training sets of varying
size $l$. We increase the size of the training sets exponentially from
$l=2^7$ to $l=2^{14}$ and for each training set we execute a full grid
search for the optimal values of $C^+$ and $C^-$ as in Experiment \#1.

Figure \ref{fig:exec_time} shows the change in the execution time as
the training size, $l$, grows exponentially for the span-rule and
$K$-fold CV, using feature dimension, $d$, equal to 20, 40 and
80. Each point on the diagrams represents the average execution time
of the corresponding hyperparameter selection method for $1089$ models
trained on a full parameter grid search. Similarly, Figure
\ref{fig:nof_SVs} shows the average number of in-bound support vectors
and the average number of total (in-bound and bounded) support vectors
for the corresponding points.

\smallskip

The experimental results can be interpreted from the following:

\begin{itemize}

\item For small training sizes $l$, the $K$-fold CV is marginally
  faster than span-rule. Also, note that for small $l$ the execution
  time of $K$-fold CV is practically unaffected by the dimension of
  the feature space, $d$, due to the fact that most kernel products
  can be stored in solver's cache.

\item Given feature dimensionality $d$, the span-rule is significantly
  more efficient than $K$-fold CV for larger training set sizes (for
  this data set, $l > 2^{12}$). As training set size increases,
  solving the $K$ weighted SVM models becomes computationally harder
  since it depends directly on $l$. On the other hand, the size of the
  quadratic problems solved by the span-rule remains small since it
  depends only on $n^*$. Thus, for large training sets, solving $n$
  small quadratic problems becomes computationally more attractive
  than solving $K$ large quadratic problems.

\item For large training sizes $l$, the execution time of $K$-fold CV
  becomes quickly inefficient as the dimensionality $d$
  increases. This is due to the fact that computationally expensive
  kernel products that do not fit in the cache need to be
  re-calculated. On the other hand, the inner products required for
  computing the value of the span, $S_p$, are calculated only once due
  to the significantly smaller size of the quadratic problems.

\end{itemize}

In conclusion, the results of Experiments \#5 and \#6 demonstrate that
the efficiency of span-rule compared to $K$-fold CV increases as: (a)
the dimensionality of the feature space increases; and (b) the size of
the training set increases, given that the number of in-bound support
vectors, $n^*$, remains relatively small.

\subsection{Summary of experimental findings}
\label{subsec:exp_summary}
In this section, we experimentally evaluated four hyperparameter
selection methods applicable to weighted SVM: the span-rule of
Corollary 1, the span bound of Theorem 1, the $K$-fold CV (with $K=5$)
and the $\xialpha$ bound. We evaluated the effectiveness and the
efficiency of the hyperparameter selection methods on $14$ data sets
in two experimental settings. Furthermore, we examined how often the
existence condition of Lemma 1 holds, which is the basis for the span
error bound theory for weighted SVM.

The key experimental findings are:
\begin{itemize}
\item The span-rule is the most effective method for weighted SVM
  hyperparameter selection (Experiments \#1 and \#2). 
\item The span-rule is the best predictor for the value of the actual
  test error in the mean-square-error sense (Experiment \#3). 
\item The condition of Lemma 1 that ensures that $\Lp \neq \emptyset$
  is satisfied almost always (Experiment \#4). This experimental
  observation verifies the theoretical assumptions used to derive the
  span-rule for weighted SVM and ensures that we can safely apply it
  in practical problems.
\item The span-rule is a fairly efficient method for hyperparameter
  selection especially for problems with large training sets and
  high-dimensional feature spaces (Experiments \#5 and \#6).
\end{itemize}

\section{Conclusions}
\label{sec:conclusions}
In this work, we presented the extension of the span error bound
theory of \cite{vapnik2000bounds} to weighted SVM.  The ground for
this extension is the proof for the necessary and sufficient condition
for defining the span of a support vector in a weighted SVM model
(Lemma 1). Building on this, we prove the span bound (Theorem 1) and
the span-rule (Corollary 1) for weighted SVM algorithm. Furthermore,
we prove that, under the assumption that the support vectors do not
change, the span is defined for all support vectors of a weighted SVM
solution (Theorem 2).

The extended theory enables new and effective tools for weighted SVM
hyperparameter selection. We experimentally evaluate the presented
span bound for weighted SVM, the presented span-rule for weighted SVM,
the $K$-fold cross-validation (with $K=5$) and the $\xialpha$ bound on
$14$ standard benchmark data sets and data sets with instance
importance scores.

The experiments demonstrate the practical value of the span error
bound theory for weighted SVM. Using the span-rule we are able to
efficienlty select training parameters that lead to small testing
error. Furthermore, using the span-rule we can accurately predict the
value of the actual test error in the mean-square-error
sense. Compared to the other methods, the span-rule demonstrates a
considerable improvement over the $K$-fold cross-validation and
performs significantly better than the $\xialpha$ bound.

Additional experiments analyzing several thousands of weighted SVM
models trained on all data sets confirmed that the condition of Lemma
1 holds true virtually for all support vectors. This finding ensures
that we can safely apply the span-rule for weighted SVM in practical
problems.

Apart from the practical applications of the span-rule for hyperparameter
selection tasks, the presented work allows further theoretical
investigation for the weighted SVM algorithm. Most importantly, new
theoretical tools are enabled---based on the geometrical concept of
the span---for understanding the effect of the instance weights and
their role in the generalization performance of the weighted SVM
algorithm.

\newpage

\section*{Appendix A. Proofs}
\section*{A.1 ~  Proof of Lemma 1}
In the first part of the proof, we show the sufficient condition for
the existence of $\Lp$ for an in-bound support vector $\x_p$ of a
weighted SVM solution. In the second part, we prove that the condition
is also necessary.

{\it Sufficiency:} Based on the proof of \cite{vapnik2000bounds} for
standard SVM in the non-separable case, let the set

\begin{equation}
\label{eq:lemma1_LpPlus}
\Lp^+ = \left\{\x ~:~ \x = \sumUBNoP \ll_i \x_i \in \Lp, \text{ and } 
       \ll_i \geq 0 ~~ \forall i \neq p
\right\}
\end{equation}

Because $\Lp^+ \subseteq \Lp$, proving that $\Lp^+ \neq \emptyset$
under the Lemma's condition also implies that the same condition
suffices for $\Lp \neq \emptyset$.

In view of \eqref{eq:SVMLambdaSet}, $\Lp^+$ contains exactly the
linear combinations $\sumUBNoP \ll_i \x_i$ that satisfy:
\begin{equation}
\label{eq:lemma1_4}
\sumUBNoP \ll_i  = 1
\end{equation}

\begin{equation}
\label{eq:lemma1_adelo_1}
0 \leq \a^0_i + y_i y_p \a^0_p \ll_i \leq C_i, ~~~
\forall i \neq p, ~~ i=1,\dots,n^*
\end{equation}

\begin{equation}
\label{eq:lemma1_adelo_2}
\ll_i \geq 0, ~~~
\forall i \neq p, ~~ i=1,\dots,n^*
\end{equation}

Equations \eqref{eq:lemma1_adelo_1} and \eqref{eq:lemma1_adelo_2} can
be expressed through the alternative conditions:

\begin{equation}
\label{eq:lemma1_1}
\ll_i = \mu \frac{C_i - \a_i^0}{\a_p^0}, ~~ y_i = y_p, ~~ i \neq p, ~~ i=1,\dots,n^*
\end{equation}

\begin{equation}
\label{eq:lemma1_2}
\ll_i = \mu \frac{\a_i^0}{\a_p^0}, ~~ y_i \neq y_p, ~~ i=1,\dots,n^*
\end{equation}

\begin{equation}
\label{eq:lemma1_3}
0 \leq \mu \leq 1
\end{equation}

Our target is now to show that condition of Lemma 1
(eq. \ref{eq:existence_condition}) guarantees that an appropriate $\mu$
exists that
makes \eqref{eq:lemma1_4}, \eqref{eq:lemma1_1}, \eqref{eq:lemma1_2}, \eqref{eq:lemma1_3}
hold simultaneously true.

Substituting equations \eqref{eq:lemma1_1} and \eqref{eq:lemma1_2}
into \eqref{eq:lemma1_4} gives
\begin{equation}
\label{eq:lemma1_5}
1 = \frac{\mu}{\a_p^0} \left(
\sumUBNoPpos(C_i - \a_i^0) + \sumUBneg \a_i^0
\right)
\end{equation}

Let $\D$ be the positive quantity:
\begin{align}
\D &= \sumUBpos ( C_i - \a_i^0 ) + \sumUBneg \a_i^0  \nonumber \\
\label{eq:lemma1_7}
   &= -y_p \sumUB y_i \a_i^0 + \sumUBpos C_i
\end{align}

Using $\D$ we can write equation \eqref{eq:lemma1_5}
as
\begin{equation}
\label{eq:lemma1_9}
1 = \frac{\mu}{\a_p^0}(\D - (C_p - \a_p^0))
\end{equation}

Hence, equation \eqref{eq:lemma1_4} is satisfied for
\begin{equation}
\label{eq:lemma1_mu}
\mu = \frac{a_p^0}{\D-C_p + a_p^0}
\end{equation}

It remains to show that this value of $\mu$ also satisfies equation
\eqref{eq:lemma1_3}.
Indeed, if the condition \eqref{eq:existence_condition} of Lemma 1
holds true for an in-bound support vector ($p = 1, \dots, n^*$), then

\begin{equation*}
\sumUBNoPpos C_i \geq - y_p \sumB y_i C_i
\end{equation*}

and the following holds true

\begin{align}
\D - C_p &= -y_p \sumUB y_i \a_i^0 + \sumUBNoPpos C_i \nonumber \\
   &\geq -y_p \sumUB y_i \a_i^0 - y_p \sumB y_i C_i \nonumber \\
   &= -y_p \left(\sumUB y_i \a_i^0 + \sumB y_i C_i\right) \label{eq:D_ineq}
\end{align}

From the constraint \eqref{eq:fsvm_dual_con2} of weighted SVM dual
problem, it is
\begin{equation}
\label{eq:constrain_trick}
\sumN y_i \a_i^0 = \sumUB y_i \a_i^0 + \sumB y_i C_i = 0
\end{equation}

Substituting \eqref{eq:constrain_trick} into \eqref{eq:D_ineq} yields
\begin{equation}
\label{eq:lemma1_9b}
\D - C_p \geq 0 
\end{equation}

Thus, from \eqref{eq:lemma1_mu} and \eqref{eq:lemma1_9b} we can show
that the condition \eqref{eq:lemma1_3} is satisfied as well.

\vskip 0.1cm

{\it Necessity:} We will show that \eqref{eq:existence_condition} is
also a necessary condition. The box constraints of $\Lp$
(eq. \ref{eq:SVMLambdaSet}) can be written as
\begin{align}
\label{eq:lemma1_11}
0 \leq \a_i^0 + \a_p^0 \ll_i \leq C_i,~~  & y_i = y_p, ~~ i=1, \dots, n^*,
~~ i \neq p \\
\label{eq:lemma1_12}
0 \leq \a_i^0 - \a_p^0 \ll_i \leq C_i,~~ & y_i \neq y_p, ~~ i=1, \dots,
n^*
\end{align}
From the summation of the inequalities given by \eqref{eq:lemma1_11} we get
\begin{equation}
\label{eq:lemma1_13}
0 \leq \sumUBNoPpos \a_i^0 + \a_p^0 \sumUBNoPpos \ll_i \leq
\sumUBNoPpos C_i 
\end{equation}
Similarly, from the inequalities given by \eqref{eq:lemma1_12} we get
\begin{equation}
\label{eq:lemma1_14}
0 \leq \sumUBneg \a_i^0 - \a_p^0 \sumUBneg \ll_i \leq
\sumUBneg C_i 
\end{equation}

By rewriting \eqref{eq:lemma1_14} as
\begin{equation}
\label{eq:lemma1_15}
- \sumUBneg C_i \leq - \sumUBneg \a_i^0 + \a_p^0 \sumUBneg \ll_i \leq 0
\end{equation}
and taking the summation of \eqref{eq:lemma1_13}
and \eqref{eq:lemma1_15} we have
\begin{equation}
\nonumber
\label{eq:lemma1_16}
- \sumUBneg C_i \leq 
\sumUBNoPpos \a_i^0 + \a_p^0 \sumUBNoPpos \ll_i - \sumUBneg \a_i^0 + \a_p^0 \sumUBneg \ll_i  
\leq \sumUBNoPpos C_i ~\Leftrightarrow
\end{equation}
\begin{equation}
\nonumber
\label{eq:lemma1_17}
- \sumUBneg C_i \leq y_p \sumUBNoP y_i \a_i^0 + \a_p^0 \sumUBNoP \ll_i
\leq \sumUBNoPpos C_i ~\Leftrightarrow
\end{equation}
\begin{equation}
\label{eq:lemma1_19}
- \sumUBneg C_i - y_p \sumUB  y_i \a_i^0 \leq  \a_p^0  \left(\sumUBNoP \ll_i
- 1 \right) \leq \sumUBNoPpos C_i  - y_p \sumUB  y_i \a_i^0
\end{equation}

By definition (eq. \ref{eq:SVMLambdaSet}) the existence of the set
$\Lp$ requires
\begin{equation}
\label{eq:lemma1_equation_constrain}
\sumUBNoP \ll_i = 1
\end{equation}

Substituting \eqref{eq:lemma1_equation_constrain}
into \eqref{eq:lemma1_19} gives the following two necessary conditions
for $\Lp \neq \emptyset$
\begin{equation}
\label{eq:lemma1_20}
- \sumUBneg C_i - y_p \sumUB y_i \a_i^0 \leq 0
\end{equation}
\begin{equation}
\label{eq:lemma1_21}
\sumUBNoPpos C_i - y_p \sumUB y_i \a_i^0 \geq 0
\end{equation}

We can write \eqref{eq:lemma1_20} as
\begin{equation}
\nonumber
\label{eq:lemma1_23}
- \sumUBneg(C_i - \a_i) - \sumUBpos \a_i \leq 0
\end{equation}
which is always satisfied since $0 < \a_i < C_i$ for 
$i=1,\dots,n^*$. 

The inequality \eqref{eq:lemma1_21} can be written
using \eqref{eq:constrain_trick} as
\begin{equation}
\label{eq:lemma1_26}
\sumUBNoPpos C_i + y_p \sumB y_i C_i \geq 0
\end{equation}

We observe that the necessary condition \eqref{eq:lemma1_26} coincides
with the lemma's condition (eq. \ref{eq:existence_condition}) and the
lemma is proved.

\vspace{0.1in}
\noindent
{\bf Remark:} Similar to standard SVM, when $\Lp^+$ exists, and
because it is a convex combination of the in-bound support vectors
(eq. \ref{eq:lemma1_LpPlus}), it holds that: $d(\x_p, \Lp^+) \leq
D_{SV} \nonumber$, where $D_{SV}$ is the diameter of the sphere
enclosing the in-bound support vectors. Also, since $\Lp^+ \subseteq
\Lp$ we get
\begin{align}
S_p = d(\x_p, \Lp) \leq d(\x_p, \Lp^+) \leq D_{SV} \nonumber
\end{align}

\section*{A.2 ~  Proof of Theorem 1}
The instances that are not support vectors are correctly
classified by the initial model. Also, their removal from the training
set does not change the decision function.  Therefore, these
training instances do not contribute in the number of leave-one-out
errors.

Considering the $k$ in-bound support vectors that do not satisfy the
condition of Lemma 1 and the $m$ bounded support vectors as
leave-one-out errors, then
\begin{align}
\Lagr(\x_1,y_1 \dots, \x_l,y_l) &\leq \Lagr^*(\x_1,y_1, \dots,\x_l,
y_l) + k + m \label{eq:total_loo_errors}
\end{align}
where $\Lagr^*(\x_1,y_1,\dots,\x_l, y_l)$ is the number of
leave-one-out errors originating by the $n^*-k$ in-bound support
vectors with $\Lp \neq \emptyset$.

In view of Lemma 2 the following inequality holds true
\begin{align}
\Lagr^*(\x_1,y_1, \dots,\x_l, y_l) \leq \sum_{p=1}^{n^*-k} \a_p^0 S_p max(D, 1/\sqrt{C_p})
\label{eq:inbound_errors}
\end{align}

Substituting \eqref{eq:inbound_errors} into
\eqref{eq:total_loo_errors} we have:
\begin{align}
\Lagr(\x_1,y_1 \dots, \x_l,y_l) & \leq \sum_{p=1}^{n^*-k}[S_p \max(D,1/\sqrt{C_p})\a_p^0] + k + m
\\ & \leq S\sum_{p=1}^{n^*-k}[\max(D,1/\sqrt{C_p})\a_p^0] + k + m \label{eq:ineq_loo_errors}
\end{align}
where we used the value of $S$-span from \eqref{eq:SSpan}.

Dividing both sides of \eqref{eq:ineq_loo_errors} by $l$ and taking the
expectation over the ensemble of training sets of size $l$, we get
\begin{align}
\label{eq:loo_ineq}
E_{\T_l}\left[\LOOError(\T_l)\right] \leq
E_{\T_l}\left[\frac{S\sum_{p=1}^{n^*-k}[\max(D,1/\sqrt{C_p})\a_p^0] + k + m}{l}\right]
\end{align}
Finally, the theorem is proved by applying the leave-one-out theorem
(eq. \ref{eq:loo_theorem}) for the left-hand side of inequality
\eqref{eq:loo_ineq}.

\section*{A.3 ~ Proof of Theorem 2}
 
Let $\aV^0 = (\a_1^0, \dots, \a_n^0, 0, \dots, 0)$ be the initial
solution on the complete training set that maximizes the objective
functional $W(\aV)$ of the dual problem of weighted SVM
(eq. \ref{eq:fsvm_dual_obj}). 

Let $\x_p$ be a fixed support vector (either in-bound or bounded) for
which a leave-one-out model is trained. Under the assumption that the
in-bound and bounded sets of support vectors do not change, let the
vector of Lagrange multipliers of the optimal solution on the reduced
training set:
\begin{equation*}
\aV^p = (\a_1^p, \dots,
\a_{p-1}^p, 0, \a_{p+1}^p,\a_n^0, 0, \dots, 0)
\end{equation*}

{\it Proof of \( \Lp \neq \emptyset \):} We will show that $\Lp \neq
\emptyset$ by finding a fixed vector $\lV'$ that satisfies both the
linear constraint and the box constraints of \eqref{eq:SVMLambdaSet}.

Using $\aV^0$ and $\aV^p$ we
set the elements of $\lV'$ as:
\begin{equation}
\label{eq:theorem4_lambdaStar}
\ll_i' = y_i y_p \frac{\a_i^p - \a_i^0}{\a_p^0}, ~~~ i=1\dots, n^*, ~ i \neq p
\end{equation}

The box constrains of \eqref{eq:SVMLambdaSet} are satisfied for $\lV'$ since
\begin{gather*}
0 \leq \a^0_i + y_i y_p \a^0_p \ll_i' \leq C_i  ~\Leftrightarrow\\
0 \leq \a_i^0 + y_i y_p \a^0_p y_i y_p \frac{\a_i^p - \a_i^0}{\a_p^0} \leq C_i ~\Leftrightarrow \\
0 \leq \a_i^p \leq C_i
\end{gather*}
which holds true.

For $\lV'$, the left-hand side of the linear constraint of
\eqref{eq:SVMLambdaSet} can be written using
\eqref{eq:theorem4_lambdaStar} as
\begin{gather}
\label{eq:theorem4_0}
\sumUBNoP \ll_i' = 
\frac{y_p}{\a_p^0} \left( \sumUBNoP y_i \a_i^p - \sumUBNoP y_i \a_i^0 \right)
\end{gather}

From the constraint \eqref{eq:fsvm_dual_con2} of dual problem, and
considering that $\x_p$ can be either in-bound or bounded ($0
< \a_p^0 \leq C_p$), the following holds for the initial solution
$\aV^0$
\begin{gather}
\sumN y_i \a_i^0 = \sumUB y_i \a_i^0 + \sumB y_i C_i = 0 ~\Leftrightarrow \nonumber \\
\label{eq:theorem4_1}
\sumUBNoP y_i \a_i^0 + \sumBNoP y_i C_i + y_p \a_p^0 = 0
\end{gather}

Similarly, from \eqref{eq:fsvm_dual_con2} for the leave-one-out
solution $\aV^p$ (where $\a_p^p = 0$) it is
\begin{gather}
\label{eq:theorem4_2}
\sumUBNoP y_i \a_i^p + \sumBNoP y_i C_i = 0
\end{gather}

Using \eqref{eq:theorem4_1} and \eqref{eq:theorem4_2} we can write
\eqref{eq:theorem4_0} as
\begin{align*}
\sumUBNoP \ll_i' = 
\frac{y_p}{\a_p^0} 
\left( -\sumBNoP y_i C_i + \sumBNoP y_i C_i + y_p \a_p^0  \right) 
= \frac{y_p}{\a_p^0} \cdot y_p \a_p^0 = 1
\end{align*}

Hence, the linear combination $\lV' = \sumUBNoP \ll_i' \x_i$ belongs in
$\Lp$. Thus, the set $\Lp$ is non-empty for any support vector $\x_p$
under theorem's assumption.

\vskip 0.3cm

{\it Proof of theorem's equality:} This part of the proof is omitted
due to its similarity to the proof of Theorem 2.3 of
\cite{vapnik2000bounds}.

\newpage

\bibliography{refs}

\end{document}